%% file: main.tex
\pdfoutput=1

\documentclass{article}

\PassOptionsToPackage{numbers, compress}{natbib}
\usepackage[preprint]{neurips_2026}

\usepackage[utf8]{inputenc} 
\usepackage[T1]{fontenc}    
\usepackage{hyperref}       
\usepackage{url}            
\usepackage{booktabs}       
\usepackage{amsmath}        
\usepackage{amsfonts}       
\usepackage{amssymb}        
\usepackage{nicefrac}       
\usepackage{microtype}      
\usepackage{xcolor}         
\usepackage{graphicx}       
\usepackage{subcaption}     
\usepackage{enumitem}       
\usepackage[normalem]{ulem} 
\usepackage{tikz}           
\usetikzlibrary{positioning, calc, fit, backgrounds,
                shapes.geometric, arrows.meta, decorations.pathreplacing,
                patterns}
\usepackage{fontawesome5}   
\usepackage{caption}
\definecolor{encFill}{RGB}{220, 233, 244}      
\definecolor{encBorder}{RGB}{55, 140, 185}     
\definecolor{encBand}{RGB}{240, 246, 251}      
\definecolor{decFill}{RGB}{252, 232, 218}      
\definecolor{decBorder}{RGB}{225, 145, 110}    
\definecolor{decBand}{RGB}{254, 247, 242}      
\definecolor{guideFill}{RGB}{248, 234, 200}    
\definecolor{guideBorder}{RGB}{190, 150, 70}   
\definecolor{guideBand}{RGB}{253, 247, 228}    
\definecolor{frozenFill}{RGB}{235, 237, 240}   
\definecolor{frozenBorder}{RGB}{140, 150, 160} 
\definecolor{flowArrow}{RGB}{70, 75, 90}       
\definecolor{imgFill}{RGB}{246, 244, 240}      
\definecolor{imgBorder}{RGB}{180, 170, 160}    
\definecolor{oursAccent}{RGB}{55, 140, 185}    
\colorlet{latentFill}{encFill}
\colorlet{latentBorder}{encBorder}
\colorlet{condArrow}{encBorder}

\input{custom_commands}
\usepackage{dsfont}
\usepackage{comment}
\usepackage{amsthm}
\usepackage[table]{xcolor}
\usepackage{multirow}

\addeditor{antoine}{AG}{0.0, 0.5, 0.0}
\addeditor{shu}{SN}{0.0, 0.7, 0.7}
\addeditor{nicolas}{ND}{0.0, 0.0, 0.8}
\addeditor{jiahui}{JL}{0.9, 0.4, 0.}
\addeditor{ko}{KN}{1.0, 0.0, 1.0}
\addeditor{ak}{AK}{0.75, 0.6, 0.3}
\addeditor{todo}{TD}{1.0, 0.0, 0.0}
\showeditsfalse

\title{Surflo: Consistent 3D Surface Flow Model\\ with Global State}

\author{%
  \textbf{Antoine Gu\'edon}$^{1,\ast}$ \quad
  \textbf{Shu Nakamura}$^{2,\ast}$ \quad
  \textbf{Nicolas Dufour}$^{3,\ast}$ \quad
  \textbf{Jiahui Lei}$^{4}$ \\[0.4em]
  \textbf{Ko Nishino}$^{2}$ \quad
  \textbf{Angjoo Kanazawa}$^{4}$ \\[0.4em]
  {\small $^{\ast}$Equal contribution.} \\[0.4em]
  $^{1}$LIX, \'Ecole polytechnique \quad
  $^{2}$Kyoto University \quad
  $^{3}$Kyutai \quad
  $^{4}$UC Berkeley \\[0.6em]
  \texttt{antoine.guedon@enpc.fr} \\[0.3em]
  \href{https://anttwo.github.io/surflo}%
    {\textcolor{oursAccent}{\texttt{https://anttwo.github.io/surflo}}}
}

\begin{document}

\maketitle

\input{figures/teaser}
\input{sections/0_abstract}

\input{sections/1_intro}
\input{sections/3_method}
\input{sections/4_experiments}
\input{sections/2_related_work_short}
\input{sections/5_conclusion}


\bibliographystyle{plainnat}
\bibliography{main}


\newpage
\appendix
\input{sections/X_appendix}



\end{document}

%% file: custom_commands.tex
\usepackage{dsfont}
\usepackage{etoolbox}
\usepackage{color}
\usepackage{xspace}

\makeatletter
\DeclareRobustCommand*\cal{\@fontswitch\relax\mathcal}
\makeatother

\newif\ifshowedits
\showeditstrue

\newcommand{\addeditor}[3]{%
  \definecolor{#1color}{rgb}{#3}
  \expandafter\newcommand\csname #1\endcsname[1]{%
  \ifshowedits
    {\color{#1color} ##1}%
  \else
    {##1}%
  \fi
  }%
  \expandafter\newcommand\csname #1rmk\endcsname[1]{%
  \ifshowedits
    {\color{#1color} {\bf [#2: ##1]}}
  \fi
  }%
  \expandafter\newcommand\csname #1rpl\endcsname[2]{%
  \ifshowedits
    {\color{#1color} ##1 \sout{##2}}
  \else
    {##1}
  \fi
  }%
}

\newcommand{\createtextvar}[1]{
  \expandafter\newcommand\csname #1\endcsname{%
  {\text{#1}}
}%
}
\newcommand{\mycomment}[1]{}

\newcommand{\bn}{{\bf n}}

\newcommand{\bp}{{\bf p}}

\newcommand{\bx}{{\bf x}}

\newcommand{\bz}{{\bf z}}

\newcommand{\bC}{{\bf C}}

\newcommand{\bI}{{\bf I}}

\newcommand{\bT}{{\bf T}}

\newcommand{\IR}{{\mathds{R}}}

\newcommand{\dxdt}{{\frac{d\bx_t}{dt}}}

\newcommand{\methodname}{Surflo\xspace}

\usepackage{colortbl}

\definecolor{rankFirst}{rgb}{1.0, 0.70, 0.70}   
\definecolor{rankSecond}{rgb}{1.0, 0.85, 0.70}  
\definecolor{rankThird}{rgb}{1.0, 1.00, 0.70}   

\definecolor{tablethree}{rgb}{0.9, 0.9, 1}
\definecolor{tabletwo}{rgb}{0.8, 0.8, 1}
\definecolor{tableone}{rgb}{0.7, 0.7, 1}

\definecolor{grey}{rgb}{0.7, 0.7, 0.7}

\newcommand{\best}{\cellcolor{tableone}}
\newcommand{\sbest}{\cellcolor{tabletwo}}
\newcommand{\tbest}{\cellcolor{tablethree}}

\definecolor{mygrey}{rgb}{.5,.5,.5}
\newcommand{\grey}{\color{mygrey}}

%% file: figures/teaser.tex
\begin{figure}[!ht]
\vspace{-24pt}
  \centering
  \includegraphics[width=\linewidth]{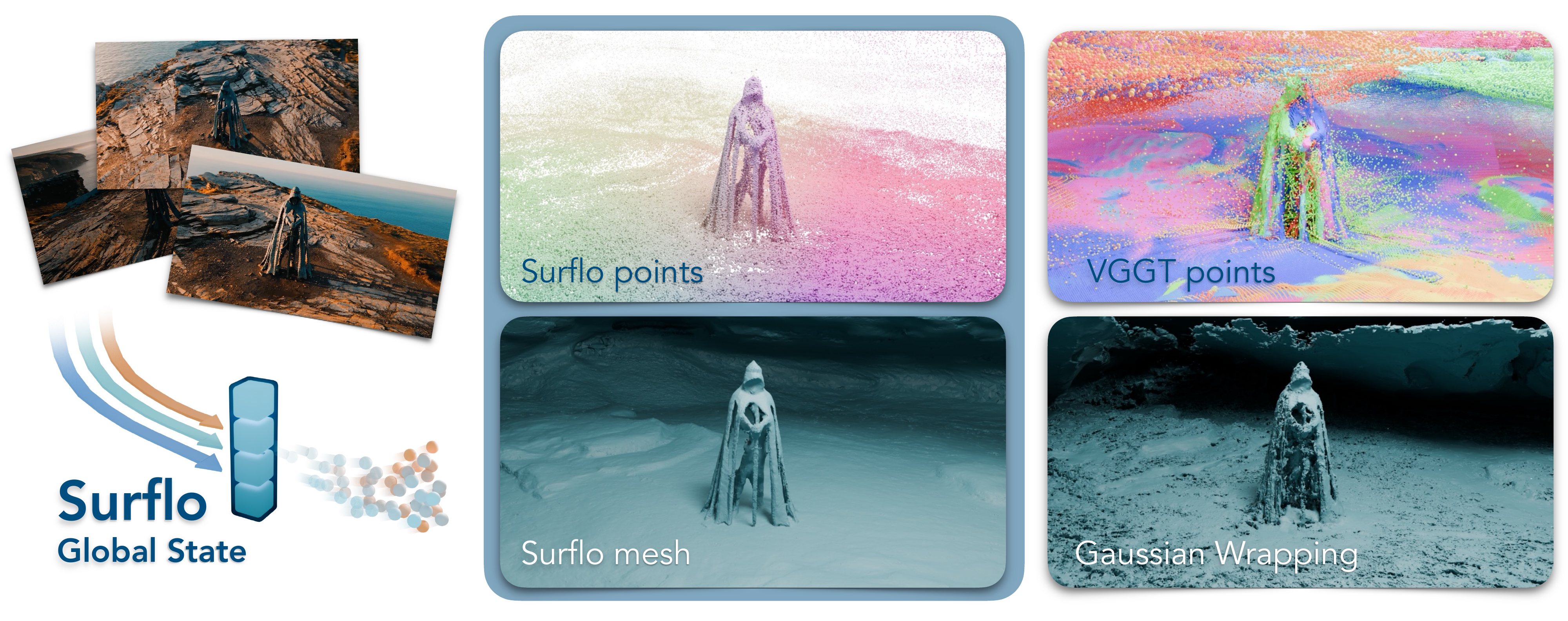}
  \caption{
   \textbf{\methodname turns a handful of unposed RGB views into a detailed 3D surface.} From a variable number of input views \emph{(left)}, \methodname encodes the scene into a fixed-size latent and decodes it via flow matching into an arbitrary number of oriented surface points, yielding a clean mesh  \emph{(middle)}. Per-view feed-forward models~\citep{wang2025vggt,lin2025depth} instead emit overlapping pointmaps \emph{(right)} that complicate mesh extraction. \methodname exceeds them in quality, while running an order of magnitude faster than state-of-the-art optimization-based methods like Gaussian Wrapping~\cite{gomez2026gaussian}, which require hundreds of views. All results are from the same 16 unposed input images, a sparse set of screenshots from an in-the-wild video of the \emph{Gallos} sculpture at Tintagel Castle, UK.
   }
    
  \label{fig:teaser}
\end{figure}

%% file: sections/0_abstract.tex
\begin{abstract}
Geometry is invariant to viewpoint, which makes any collection of images a redundant encoding of a single 3D state. Existing feed-forward reconstruction models fail to exploit this: per-view methods emit overlapping, unaligned pointmaps that grow linearly with input count, while global-latent methods commit to a fixed, low-resolution output. We introduce \methodname, which compresses a variable number of unposed RGB views into $K$ latent tokens---one global state---and decodes oriented 3D surface points by independently transporting them from noise onto the surface via flow matching. This frees the output from any fixed grid or token budget: the same latent yields from a few thousand to a million points in a single forward pass. To suppress the local inconsistencies inherent to independent per-point decoding, an inference-time guidance term correlates nearby points by injecting a photometric gradient during ODE integration. \methodname matches or surpasses feed-forward baselines on surface metrics, runs an order of magnitude faster than optimization-based methods that require hundreds of views, and is the only feed-forward approach to combine a global latent with arbitrary-resolution decoding.
\end{abstract}

%% file: sections/1_intro.tex
\section{Introduction}
\label{sec:intro}

{In the spirit of Klein's Erlangen program~\cite{klein1893}, 3D geometry is precisely what remains invariant under the viewing transformation. It follows that a collection of images of a scene is a redundant encoding of its geometry: the same underlying 3D state is re-projected into every view, modulo the camera. As more views are gathered, the raw pixel data grows linearly while the geometric content does not---additional views serve only to resolve ambiguities and refine a shared, global state. For downstream applications such as real2sim for robotics~\cite{allshire2025videomimic} and agent navigation~\cite{xia2026habitatgs}, it is this global geometry that is the target; the images are merely the instruments used to recover it.

This observation has a direct architectural consequence: if many views encode one geometry, the right intermediate representation is a single global state, not a per-view or per-pixel one. We present \textbf{\methodname}, a feed-forward model built around this principle. Given a variable number of unposed RGB views, \methodname compresses them into a single compact and consistent global representation of the 3D scene. Rather than being bound to a fixed spatial grid or a rigid token budget, the underlying geometry can be queried from this global state by sampling an arbitrary number of oriented surface points---from a few thousand to a million per scene. The resulting output is global, consistent across views, and readily converted into a mesh.
}

Existing feed-forward methods fall short of this principle along two distinct axes. \emph{Per-view} models such as VGGT~\cite{wang2025vggt,wang2024dust3r,lin2025depth} predict a separate pointmap per input image, preserving the very view-wise redundancy a global representation should compress away: their token count grows linearly with views, and the pointmaps overlap heavily without exact alignment, resisting fusion into a clean mesh (Figure~\ref{fig:teaser}). 
The few methods that summarize the scene in a single latent pay a different price: \citet{chen2026nova3r} compresses few-view inputs into a fixed-size code but reads it out as a small, fixed set of $10K$ points, and \citet{wang2025cut3r} maintains a persistent state decoded as per-view RGB rather than a global geometric object.
D4RT~\citep{zhang2026efficientlyd4rt} also adopts a scene representation queried for independent points, but operates on dense videos and regresses positions of \emph{pixel-anchored} queries rather than generatively sampling free oriented surface points from sparse views as \methodname does.

Building a representation that is both globally compressed and geometrically queryable requires three coupled ingredients: (1) compression of an unbounded number of multi-view tokens into a fixed-size representation without loss of scene-specific detail, (2) decoding of that global representation into an arbitrary number of surface points, and (3) consistency across the decoded geometry. First, the encoder of \methodname builds on a frozen VGGT~\citep{wang2025vggt} backbone for its strong geometric features: patch tokens from several intermediate layers are enriched with a 3D positional encoding of their unprojected position and compressed into $K$ latent tokens. 
The decoder then generates oriented surface points via flow matching~\citep{lipman2023flow}, but unlike standard generation on fixed grids it denoises each point independently: conditioned on the global latent, it predicts a velocity that transports an individual query point from a source distribution onto the surface. The same independence, however, has a cost---nothing forces independently sampled points to land on the same surface, so ambiguity in occluded regions can produce local inconsistencies. We resolve this with a guidance mechanism, reminiscent of loss-conditioned diffusion guidance~\citep{bansal2023universal}, that correlates nearby points at inference time: injecting the gradient of a photometric and depth loss into the predicted velocity aligns the output with the visible images and suppresses noisy outliers.

Across 8 3D reconstruction benchmarks, \methodname matches or surpasses recent feed-forward baselines on surface metrics while being the only feed-forward method that decodes geometry at arbitrary resolution from a single shared latent. The contributions of this work can be summarized as follows:
\begin{itemize}[topsep=2pt,itemsep=0pt,leftmargin=1.2em]
  \item A feed-forward encoder, built on a frozen VGGT backbone~\citep{wang2025vggt}, that compresses a variable number of unposed RGB views into a single global state independent of view count.
  \item A flow-matching decoder that reads out this latent as an oriented surface by transporting oriented points in $\IR^3\times\mathbb{S}^2$ independently, enabling arbitrary output resolution from the same shared latent.
  \item A rendering-based guidance scheme during ODE integration that correlates nearby points and aligns the decoded surface with the input images, mitigating the outliers typical of independent per-point decoding.
  \item As an auxiliary contribution, a meshed version of the DL3DV~\citep{ling2024dl3dv} dataset: each of the \textasciitilde10.5K scenes is enriched with a watertight surface mesh and \textasciitilde$10^7$ oriented points obtained via Gaussian Wrapping~\citep{gomez2026gaussian}. To our knowledge, this is the first real scene-level watertight mesh dataset at this scale, and we will release the data as well as the code upon publication.
\end{itemize}

%% file: sections/3_method.tex
\section{Method}
\label{sec:method}

\input{figures/pipeline2}

Our goal is to recover scene geometry from a variable number of input images at a variable output resolution, with both chosen freely at inference. The design of \methodname follows directly from this goal. To support an arbitrary number of input views $N$, the conditioning representation must have a size independent of $N$. Recent multi-view foundation models~\citep{wang2025vggt, wang2026pi3, lin2025depth} integrate observations across views and produce geometry-aware features, but their token count grows linearly with $N$. We therefore compress these features into a \emph{fixed-size} latent $\bz $.
To support an arbitrary number of output points $P$, we decode each query point \emph{independently}: a per-point estimator maps each query $\bx\in\IR^3\times\mathbb{S}^2$ to a surface sample (3D coordinates and normal). Decoding $P$ points thus reduces to $P$ independent, batched forward passes. To ensure that independently sampled queries fall onto a coherent surface, we introduce an inference-time guidance term that correlates nearby points. Putting these together, \methodname operates as follows (Figure~\ref{fig:pipeline}):
\begin{enumerate}[topsep=2pt,itemsep=1pt,leftmargin=1.4em]
  \item An \emph{encoder} $E_\phi$ that maps $\{\bI_n\}_{n=1}^{N}$ to a fixed-size latent $\bz \in \IR^{K\times D}$, independent of $N$.
  \item A \emph{flow-matching decoder} $v_\theta$ that, given $\bz$, a time $t\in[0,1]$, and a query $\bx_t\in\IR^3\times\mathbb{S}^2$, predicts a velocity transporting $\bx_t$ toward the scene surface $\mathcal{S}\subset\IR^3$ and its normals.
  \item A \emph{guided ODE solver} that, at inference, integrates $v_\theta$ together with a rendering-based coupling term to produce a coherent surface.
\end{enumerate}

\subsection{Encoder: From Images to a Fixed-size Global State}
\label{sec:encoder}
\noindent\textbf{Frozen VGGT backbone.}
We rely on a frozen foundation model, namely VGGT~\citep{wang2025vggt}, as the backbone for its strong geometry-aware features. 
For each input view $\bI_n$, VGGT produces a stack of patch tokens together with a single camera token. Following the procedure of \citet{wang2025vggt}, we extract and concatenate patch tokens from four layers $\ell\in\{4,11,17,23\}$, then project them to a working dimension $D$. This yields a per-view set of tokens $\bT_n \in \IR^{4N_p \times D}$, where $N_p$ is the number of patches per image. We apply a similar projection to the per-view camera tokens $\bC_n\in\IR^{4\times D}$.

\noindent\textbf{3D positional encoding.}
We augment each patch token with a 3D positional encoding. We read off the 3D coordinates of the patch centers in the corresponding VGGT pointmaps, yielding $\bp_n\in\IR^{N_p\times 3}$; we then map them to Fourier features $\gamma(\bp_n) \in \IR^{N_p\times D}$~\citep{tancik2020fourier}, and add the result to the patch tokens.
A similar encoding $\gamma$ is reused by the decoder for query points.
This shared encoding is the only mechanism that gives the decoder a notion of \emph{where} in 3D space a query lies relative to the encoded scene. Without it, the latent $\bz$ stores observations from VGGT's view-aggregated coordinate frame, while decoder queries arrive as raw coordinates with no a priori relationship to that frame; the alignment would have to be learned implicitly, across whatever coordinate system VGGT produces for each scene. With shared $\gamma$, a query point near a VGGT patch shares similar Fourier features with it, so cross-attention localizes relevant scene information by spatial proximity rather than learned association---making coordinate-frame robustness structural rather than something the decoder must absorb during training.

\noindent\textbf{Compression.}
We compress the union of $N\times4N_p$ position-encoded patch tokens $\{\bT_n + \gamma(\bp_n)\}_{n=1}^{N}$  with a Perceiver-style cross-attention module~\citep{jaegle2021perceiver,jaegle2022perceiver}. A fixed set of $K (\ll N\times 4N_p) $ learned latent queries $\bz_p^{(0)} \in \IR^{K\times D}$ cross-attends to the patch tokens, followed by $L_\mathrm{s}$ self-attention blocks:
\begin{equation}
  \bz_p^{(l+1)} \,=\, \mathrm{SelfAttn}^{L_\mathrm{s}}\ \!\bigl(\mathrm{CrossAttn}\bigl(\bz_p^{(l)},\, \{\bT_n + \gamma(\bp_n)\}_n\bigr)\bigr) \;\in\; \IR^{K\times D} \> .
\end{equation}
This process is iterated $L_\mathrm{e}$ times to obtain the final latent $\bz_p:=\bz_p^{(L_\mathrm{e})}$.
Camera tokens $\{\bC_n\}_n$ are processed by a similar but lighter Perceiver into a single latent $\bz_c \in \IR^{1\times D}$ encoding additional information about the current world space of the scene, and concatenated into $\bz := [\bz_p, \bz_c]$.

\subsection{Flow-Matching Decoder with Independent Query Points}
\label{sec:decoder}

\noindent\textbf{Per-point flow.}
We treat surface decoding as transporting individual query points in $\IR^3\times\mathbb{S}^2$ from a source distribution $p_0$ to the surface distribution $p_1$, conditioned on the latent representation $\bz$. 
We adopt the flow-matching formulation of \citet{lipman2023flow}: at training time, we sample a target $\bx_1\sim p_1$, a source $\bx_0\sim p_0$, a time $t\sim\mathrm{LogitNormal}(1,1.6)$, and form the linear interpolant $\bx_t = (1-t)\bx_0 + t\,\bx_1$ with conditional velocity $u(\bx_t,t) = \bx_1 - \bx_0$. 

\noindent\textbf{Decoder architecture.}
The decoder $v_\theta$ is an $L$-layered transformer that ingests a Fourier-encoded query point $\gamma(\bx_t)$ and predicts a velocity through cross-attention to the latent $\bz_p$ and an MLP.
The prediction of the velocity is also conditioned on the time and camera token $\left[\tau(t), \bz_c\right]$ through Ada-LN~\cite{peebles2023scalable} blocks before each cross-attention and MLP, where $\tau$ is a sinusoidal time embedding, allowing the decoder to adjust its prediction to the current time and world space of the scene.

\noindent\textbf{Training objective.}
The training objective is the standard flow-matching loss
\begin{equation}
  \mathcal{L}_{\text{FM}}(\theta,\phi) \;=\; \mathbb{E}_{t,\bx_0,\bx_1,\,\{\bI_n\}_n}\;\bigl\|\, v_\theta\!\bigl(\bx_t,\, t\ |\ \bz\bigr) \,-\, (\bx_1-\bx_0) \,\bigr\|_2^{2}\, \> .
  \label{eq:fm_loss}
\end{equation}
The loss decomposes over independent query points, allowing us to train on thousands of points per scene while keeping a single shared latent in memory. Parameters $\theta$ and $\phi$ are learnt jointly, as the decoder and encoder are trained together in an end-to-end fashion.

\input{figures/guidance_anim}

\noindent\textbf{Source distribution.}
Each query's normal is initialized uniformly on $\mathbb{S}^2$. Initializing the 3D coordinate from $\mathcal{N}(0,\bI_3)$, however, wastes model capacity on empty space and slows convergence. We instead sample from a mixture of Gaussians centered on perturbed VGGT pointmap samples, with noise $\sigma_s$ large enough to cover surfaces occluded in the input views (Figure~\ref{fig:guidance}). This concentrates the flow near the geometry while leaving room to fill in occluded regions.

\noindent\textbf{Inference.}
At inference, we sample $P$ query points $\{\bx_0^{(m)}\}_{m=1}^{P}$ from $p_0$ and integrate $\dxdt = v_\theta(\bx,t,\bz)$ from $t{=}0$ to $t{=}1$ with a simple Euler solver. 
Because all $P$ queries share the same latent $\bz$, the per-point cost is dominated by the relatively small decoder, and we can decode $P=10^3$ to $P=10^6$ points in a batched pass on a single GPU. 
As a result, depending on the requested number of samples, the same scene latent supports both a fast coarse preview and a dense surface (Figure~\ref{fig:multires}).

\subsection{Communication via Guidance}
\label{sec:guidance}

Per-point independence gives \methodname its flexibility, but at a cost: Equation~\eqref{eq:fm_loss} models each query in isolation, never the joint distribution over surface points. As a result, independently sampled queries are not constrained to land on the same physical surface, producing scattered points between surfaces and drift from the input views. We therefore augment the deterministic flow with a \emph{communication-via-guidance} term, in the spirit of universal guidance~\citep{bansal2023universal,he2024manifold}, injecting the gradient of a differentiable-rendering~\citep{kerbl2023gaussian,zhang2024radegs,guedon2025matcha} loss into each ODE step so independently sampled queries can coordinate through a shared global signal.

Specifically, near the end of the ODE integration, for $t\geq 0.95$, we predict at each ODE step the corresponding target points $\hat\bx_1 = \bx_t + (1{-}t)\,v_\theta$ and run $M$ gradient-descent steps on a global communication loss $\mathcal{L}_{\mathrm{render}}$ that depends on the entire batch. This process couples all points and adjusts the estimated target points, yielding a guided target $\hat\bx_1^{\,\mathrm{g}}$ and the corresponding guided velocity $v_{\mathrm{g}} = (\hat\bx_1^{\,\mathrm{g}} - \bx_t)/(1-t)$ used for the actual Euler step $\bx_{t+dt} = \bx_t + dt\,v_{\mathrm{g}}$.

In practice, we use a rendering-based loss: we treat the set of predicted targets $\hat\bx_1$ as a cloud of small oriented Gaussians and \emph{render} them through the cameras recovered by the pretrained VGGT backbone to obtain images $\hat\bI_n(t)$ at the input view poses. 
Following Gaussian Wrapping~\citep{gomez2026gaussian}, we then compute a rendering loss against the input views:
\begin{equation}
  \mathcal{L}_{\text{render}}\bigl(\{\hat\bx_1\}\bigr) \;=\; \frac{1}{N}\sum_{n=1}^{N}\, \lambda \, \bigl\|\,\hat\bI_n(t) \,-\, \bI_n\,\bigr\|_1 
  + (1-\lambda) \, \text{DSSIM}(\hat\bI_n(t) \,, \bI_n) \> ,
\end{equation}
including additional regularization terms on rendered depth maps~\citep{gomez2026gaussian,guedon2025matcha}.

The updated velocities \emph{couple} the query points: nearby queries that disagree about a surface receive a shared corrective signal, which acts as a soft consistency prior. The rendering objective also encourages the points to match the input images and improves the quality of the details considerably. Optionally, we strengthen guidance with a monocular-depth expert~\citep{lin2025depth} (Figure~\ref{fig:guidance}). Monocular depth predictions are not multi-view consistent, but their relative ordering provides a strong guidance signal via depth-order regularization~\citep{guedon2025matcha,guedon2025milo}. 

The resulting oriented Gaussians are converted into a triangle mesh via Delaunay triangulation, following~\citep{gomez2026gaussian}.

\subsection{Training data: Explicit Surfaces for DL3DV}
\label{sec:data}

Scene-level 3D surface datasets of real environments do not exist at the scale we need: Tanks~\&~Temples~\citep{knapitsch2017tanks} or BlendedMVS~\citep{yao2020blendedmvs} contain only a few dozen scenes, and DL3DV~\citep{ling2024dl3dv} provides multi-view images and COLMAP poses but no surface ground truth. As an auxiliary contribution, we close this gap by enriching each of the $\sim$10.5K DL3DV scenes with a watertight mesh and an oriented point cloud. For each scene, we run Gaussian Wrapping~\citep{gomez2026gaussian} on the full set of views to extract a watertight surface, and sample $10^7$ points and normals from the mesh, discarding outliers via a visibility check against the input views. We will release this dataset and expect it to be useful for scene-level surface learning beyond \methodname.

We use this dataset as our ground-truth surface distributions $p_1$. At each training step, we first sample $12$ scenes; then, we draw $N\in[2,16]$ input views for each scene as well as a batch of $\sim8$K points uniformly on each ground-truth surface; finally, we run our model, align the ground-truth points with our predicted points, and minimize equation~\eqref{eq:fm_loss}. More details can be found in Appendix.
Rendering guidance is only applied at inference; the encoder backbone (VGGT) is kept frozen throughout.

%% file: figures/pipeline2.tex
\begin{figure}[t]
    \centering
    \includegraphics[width=\linewidth]{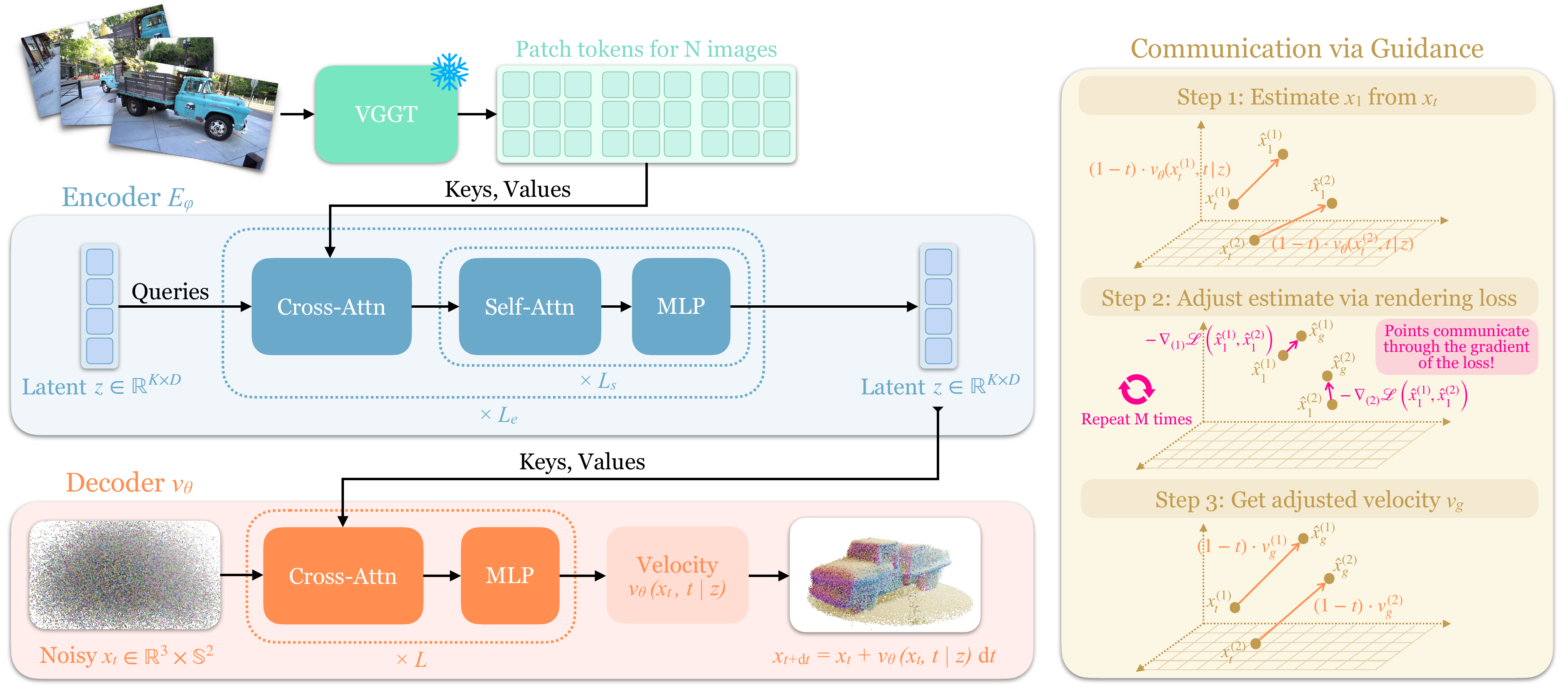}
    \caption{\textbf{Three key ingredients define \methodname.}
      \textbf{Encoder:} a frozen VGGT~\citep{wang2025vggt} maps $N$ unposed RGB views to patch tokens, which a Perceiver-style compressor with $K$ learnable queries distills into a fixed-size latent $\bz \in \IR^{K\times D}$.
      \textbf{Decoder:} each query $\bx_t\in\IR^3\times\mathbb{S}^2$ is diffused \emph{independently} via cross-attention with $\bz$ to a velocity $v_\theta$; any number of points can be integrated in parallel.
      \textbf{Guidance} (Sec.~\ref{sec:guidance}): for $t\geq 0.95$, at each step of the ODE integration, per-point velocities are coupled~\citep{bansal2023universal} via $M$ gradient steps on a global rendering loss $\mathcal{L}$.}
    \label{fig:pipeline}
  \end{figure}

%% file: figures/guidance_anim.tex
\begin{figure}[t]
  \centering
  \newlength{\guidw}\setlength{\guidw}{0.185\linewidth}
  \newlength{\guidh}\setlength{\guidh}{0.7\guidw}
  \newcommand{\guidimg}[2][\guidw]{%
    \tikz{\node[inner sep=0pt, outer sep=0pt, rounded corners=3pt, clip]
      {\includegraphics[width=#1]{#2}};}%
  }
  \newcommand{\guidph}[2][\guidw]{%
    \setlength{\fboxsep}{0pt}%
    \fcolorbox{imgBorder}{imgFill}{%
      \begin{minipage}[c][\guidh][c]%
        {\dimexpr#1-2\fboxrule\relax}\centering
        \footnotesize\itshape\color{black!50} #2
      \end{minipage}%
    }%
  }
  \setlength{\tabcolsep}{2pt}
  \renewcommand{\arraystretch}{1.15}
  \begin{tabular}{@{}ccccc@{}}
    \footnotesize\textbf{(a) Reference}                                 &
    \footnotesize (b) Source noise                                      &
    \footnotesize (c) No guidance                                       &
    \footnotesize (d) Photometric                                       &
    {\footnotesize (e) Photo + Monodepth} \\[1pt]
    \includegraphics[width=\guidw]{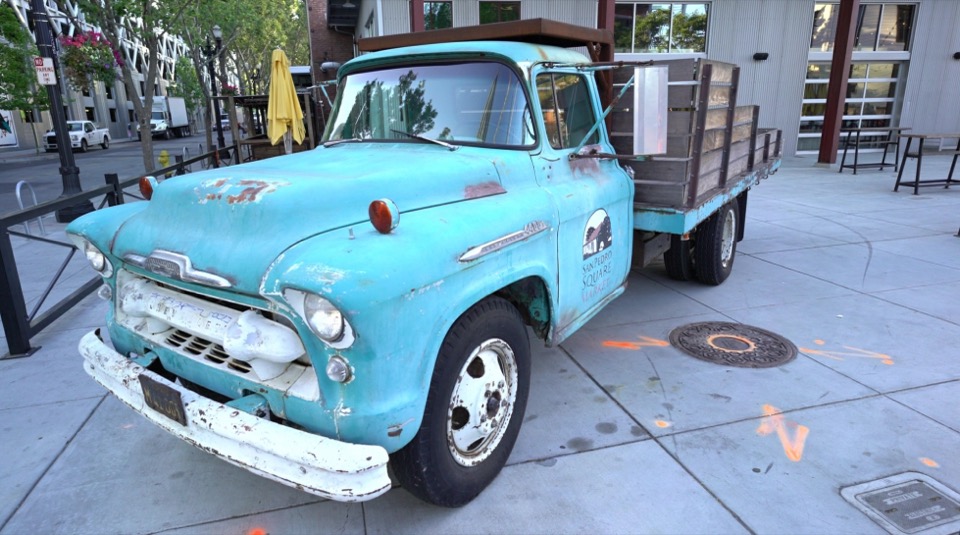} & 
    \includegraphics[width=\guidw]{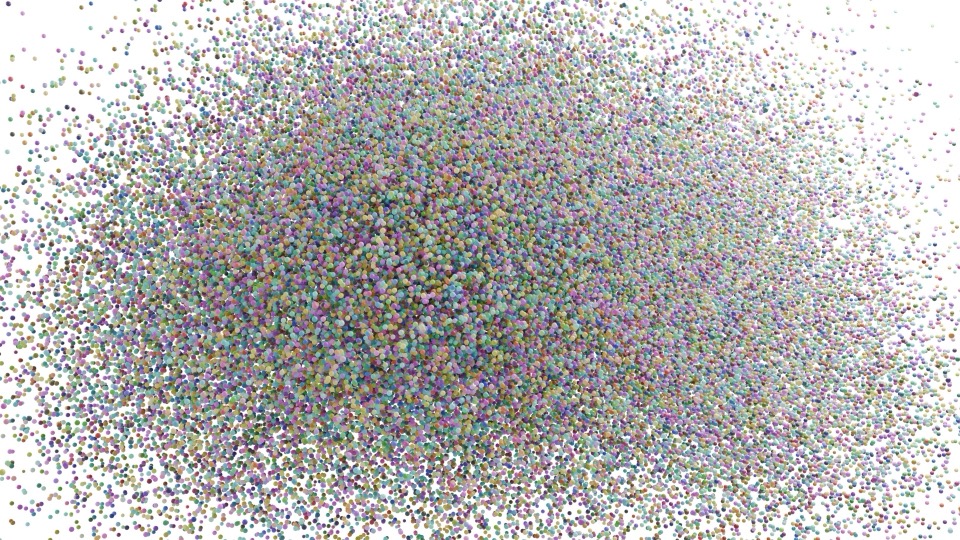}  &
      \includegraphics[width=\guidw]{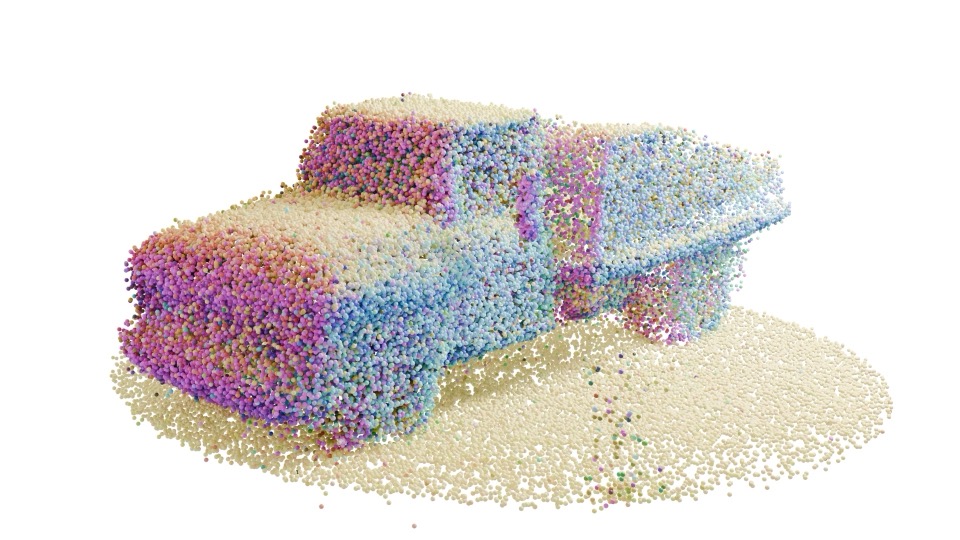}        &
      \includegraphics[width=\guidw]{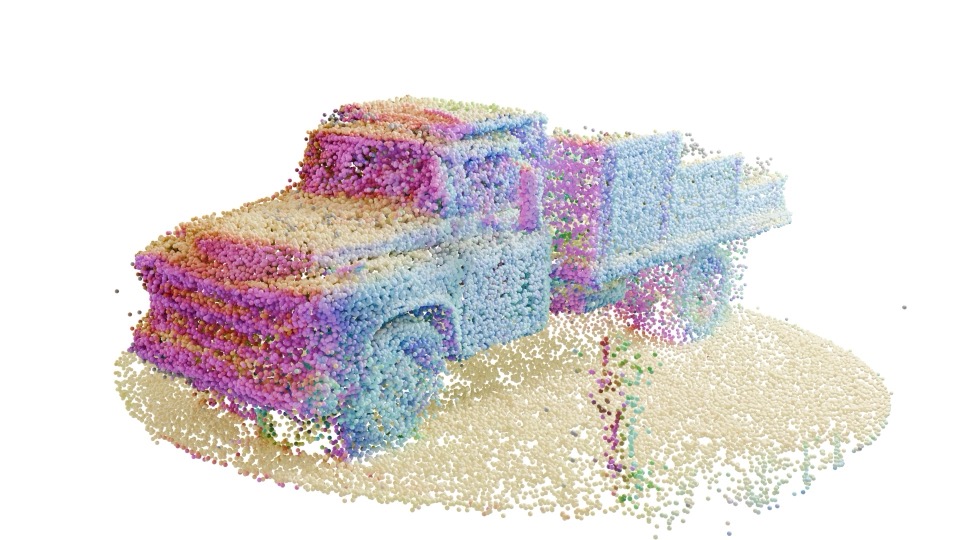}     &
      \includegraphics[width=\guidw]{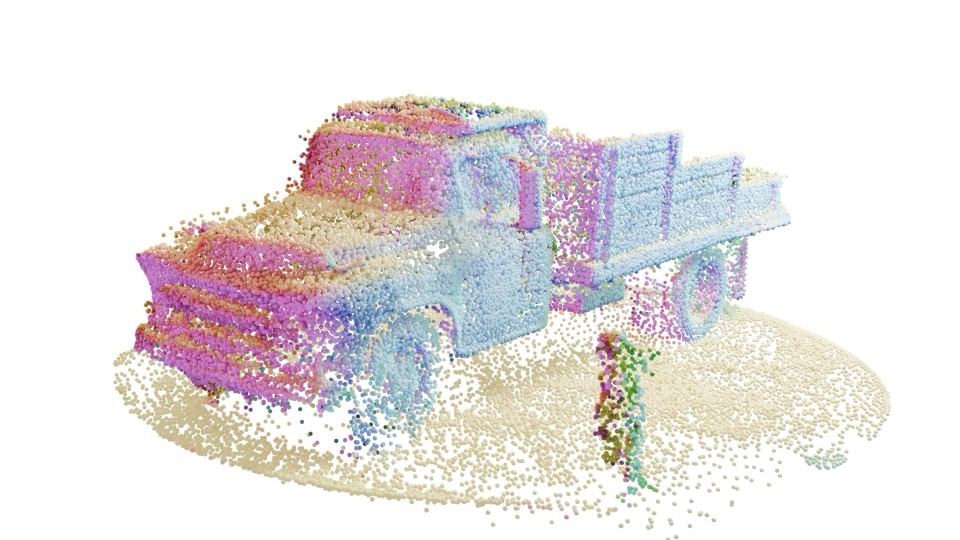} \\
  \end{tabular}
  \caption{\textbf{
  \methodname denoises 3D points into an oriented surface.
    } 
From a variable number of input views (a), \methodname denoises an arbitrary number of noisy 3D points and normals (b) into an oriented surface (c). Integrating the flow-matching velocity without guidance already achieves high accuracy but can exhibit outliers and miss fine details. 
    We therefore introduce two guidance terms. A photometric guidance term $\nabla_{\bx}\mathcal{L}_{\text{photo}}$ computed by instantiating Gaussian Splatting~\citep{kerbl2023gaussian} on the points, pulls them onto surfaces consistent with the input views and recovers finer details (d). An optional monocular depth~\citep{lin2025depth} guidance further sharpens geometry (e).
    }
  \label{fig:guidance}
\end{figure}

%% file: sections/4_experiments.tex
\section{Experiments}
\label{sec:experiments}

\input{figures/qualitative}

We evaluate \methodname on several few-view 3D reconstruction benchmarks, against representative per-view, latent and optimization-based baselines, and ablate its main design choices. As shown in Figure~\ref{fig:qualitative}, \methodname remains robust even in challenging capture conditions, including scenes with strong exposure variation across views and transparent objects, such as the semi-transparent figure in the last row.

\subsection{Setup}
\label{sec:setup}
\noindent\textbf{Datasets.}
We train on the \textasciitilde10.5K scenes of DL3DV~\citep{ling2024dl3dv} with ground-truth surfaces obtained from Gaussian Wrapping~\citep{gomez2026gaussian}, and split out a held-out test set. We evaluate on two groups of out-of-distribution benchmarks.
The first group ships with native surface ground truth: ML-Hypersim~\citep{roberts2021hypersim}, BlendedMVS~\citep{yao2020blendedmvs}, DTU~\citep{aanaes2016dtu}, and SCRREAM~\citep{jung2024scrream}.
For the second group, no whole-scene surface ground truth is available, and we generate pseudo-ground-truth surfaces from the dense view sets using the same Gaussian Wrapping pipeline we apply at training time. This group includes our DL3DV test split, Mip-NeRF~360~\citep{barron2022mip} and DeepBlending~\citep{hedman2018deep}. We also include Tanks~\&~Temples~\citep{knapitsch2017tanks}: its native ground truth covers only the foreground object of each scene, and our pseudo-GT additionally captures the background geometry. This second evaluation lets us quantify the geometry gap between feed-forward models that observe only a few input views and the best surfaces that can be extracted from a dense RGB capture.
For evaluation, each test scene is reconstructed from a fixed set of $16$ unposed input views unless stated otherwise.

\paragraph{Metrics.}
We report Chamfer Distance (CD$\downarrow$) between predicted and ground-truth point clouds (normalized by the spatial extent of the scene) and F1-score (F1$\uparrow$) at a fixed scene-relative threshold of $1\%$ of the scene diagonal, following standard practice.

\noindent\textbf{Baselines.}
We compare \methodname against three families of baselines for scene-level surface reconstruction, all of which receive the same input views (from $2$ to $16$ views) unless stated otherwise.
(i) \emph{Per-view feed-forward}: the raw pointmaps of VGGT~\citep{wang2025vggt} and Depth\-Anything-3~\citep{lin2025depth}, together with their TSDF-fused~\citep{curless1996volumetric} variants for turning per-view predictions into a single mesh. Because pointmap outputs do not represent a single coherent surface, they trade physical consistency for purely local accuracy; we therefore report them \emph{for reference only} and exclude them from the ranking.
(ii) \emph{Latent feed-forward}: NOVA3R~\citep{chen2026nova3r}, which compresses the input views into a latent but reads it out as a fixed-size point cloud. NOVA3R is evaluated with randomly sampled 2 views every time, as it is trained with 2 input views only and feeding more than 2 views systematically results in blurry reconstructions. We restrict our comparisons to methods targeting scene-level reconstruction and therefore exclude object-centric feed-forward models such as ReconViaGen~\citep{chang2025reconviagen}: their outputs cover only foreground objects and would yield uninformative scene-level metrics, as most of the surrounding geometry would be missing. 
(iii) \emph{Per-scene optimization}: 2DGS~\citep{huang20242d}, RaDe-GS~\citep{zhang2024radegs}, and Gaussian Wrapping~\citep{gomez2026gaussian} run on the same sparse views with camera parameters and initial points estimated by VGGT. These baselines highlight how strongly per-scene optimization underperforms in the few-view regime, and motivate the use of a learned prior.

\noindent\textbf{Implementation details.}
The encoder uses a frozen VGGT-1B backbone, a Perceiver compressor with $K=128$ latent tokens of dimension $D=512$, $L_s=4$ self-attention blocks per cross attention, and Gaussian Fourier features~\citep{tancik2020fourier} with $F=512$ frequencies. The decoder is a $12$-layer transformer that cross-attends to the latent for the first 6 layers. We train with AdamW, batch size $12$ scenes, $8$K query points per scene per step, on $4\times$H100 GPUs for 400K iterations. 
At inference, we use an Euler ODE solver with $150$ steps, and decode $100$K points by default, though this number can be adjusted at will; Figure~\ref{fig:multires} illustrates reconstructions with different numbers of points. Optionally, we apply photometric guidance starting from $t=0.95$, and perform $M=32$ loss updates at each ODE step.

\subsection{Results}

\input{tables/pseudo_gt_results}
\input{tables/main_results}
\noindent\textbf{Main Results.}
\label{sec:main_results}
Table~\ref{tab:main_results} reports surface metrics on the four out-of-distribution benchmarks with native surface ground truth, and Table~\ref{tab:pseudo_gt_results} reports the same metrics on DL3DV, T\&T, Mip-NeRF~360 and DeepBlending, where reference surfaces are generated by Gaussian Wrapping on the dense view sets. 
All methods receive the same $16$ input views.
The two tables tell a consistent story: per-view baselines, while accurate per-frame, accumulate misalignments when their pointmaps are merged, and their fused meshes show duplicated surfaces and eroded or floating geometry, particularly at view boundaries.
NOVA3R~\citep{chen2026nova3r} produces coarse and eroded outputs as its decoder is forced to a fixed point budget.
\methodname outperforms the best optimization-based references in absolute Chamfer at a fraction of the cost (see Efficiency in Section~\ref{sec:efficiency}), and is the only feed-forward model that also supports decoding the same latent at any chosen resolution. Figure~\ref{fig:qualitative} shows qualitative comparisons.

\input{figures/multires}

\noindent\textbf{Varying number of input views.}
\label{sec:nviews}
Beyond the $16$-view protocol used on all datasets above, we further study how \methodname behaves as the number of input views changes. Because the encoder compresses any number of views into a fixed-size latent, the same model can be evaluated across capture densities without retraining or any architectural change.
Tables~\ref{tab:views_tnt} and~\ref{tab:views_mipnerf} report this experiment on two out-of-distribution datasets, Tanks~\&~Temples and Mip-NeRF~360, as the number of unposed input views per scene varies from $2$ to $32$. 
\methodname is consistently best across all view counts, including the very sparse $2$-view regime. While \methodname benefits from additional views, the size of the latent does not grow with the number of views. 
Figure~\ref{fig:nviews} shows that even from very few views the model already produces a coherent surface; additional views progressively fill previously missing geometry and sharpen fine detail, while the decoding cost remains unchanged.

\noindent\textbf{Multi-resolution decoding.}
\label{sec:multires}
A key feature of \methodname is that the latent itself does not change with the requested output resolution. Figure~\ref{fig:multires} shows the same scene latent decoded at three resolutions, from $8$K to $128$K points, on the same GPU. The coarse output is suitable for fast collision queries or scene previews, while the dense output offers sharper details and background geometry.

\input{tables/multiview_tnt}
\input{tables/multiview_mipnerf}
\input{figures/nviews}

\noindent\textbf{Ablations.}
\label{sec:ablations}
Table~\ref{tab:ablations} ablates the three main design choices of \methodname.
\textbf{Latent size.} Increasing $K$ from $32$ to $128$ tokens consistently improves both metrics. We use $K=128$ throughout.
\textbf{Source distribution.} Replacing our noisy-VGGT-pointmap source with a pure Gaussian source degrades reconstruction, especially for large scenes.
\textbf{3D positional encoding.} Augmenting VGGT tokens with our 3D positional encoding improves both metrics.

\noindent\textbf{Photometric guidance.} Disabling the guidance term can introduce visible noisy outliers, while still resulting in high accuracy. 
Full photometric guidance is consistently best for visual quality.

\input{tables/ablations}

\noindent\textbf{Efficiency.}
\label{sec:efficiency}
We additionally report wall-clock numbers on a single H100 GPU: encoding a $16$-view scene takes a single forward pass through VGGT and the Perceiver compressor; the latent is then re-used for all subsequent queries. Decoding $10^5$ points from the cached latent takes a few seconds, dominated by the ODE solve, which is two orders of magnitude faster than per-scene optimization baselines such as 2DGS or Gaussian Wrapping. A small number of guidance steps adds modest overhead (typically from 30 seconds to 3 minutes depending on the number of guidance steps), as rendering Gaussians is inexpensive and independent of the number of views $N$.

%% file: figures/qualitative.tex
\begin{figure}[t]
  \centering
  \newlength{\qualw}\setlength{\qualw}{0.185\linewidth}
  \newlength{\qualh}\setlength{\qualh}{0.7\qualw}
  \newcommand{\qualimg}[2][\qualw]{%
    \tikz{\node[inner sep=0pt, outer sep=0pt, rounded corners=0pt, clip]
      {\includegraphics[width=#1, height=0.5625#1]{#2}};}%
  }
  \newcommand{\qualph}[2][\qualw]{%
    \setlength{\fboxsep}{0pt}%
    \fcolorbox{imgBorder}{imgFill}{%
      \begin{minipage}[c][\qualh][c]%
        {\dimexpr#1-2\fboxrule\relax}\centering
        \footnotesize\itshape\color{black!50} #2
      \end{minipage}%
    }%
  }
  \setlength{\tabcolsep}{2pt}
  \renewcommand{\arraystretch}{1.15}
  \begin{tabular}{@{}ccccc@{}}
    \footnotesize\textbf{Reference}                                    &
    \footnotesize VGGT~\citep{wang2025vggt} points                     &
    \footnotesize GW~\citep{gomez2026gaussian} mesh                    &
    {\color{oursAccent}\footnotesize\textbf{\methodname{} points}}     &
    {\color{oursAccent}\footnotesize\textbf{\methodname{} mesh}}       \\[1pt]
    \qualimg{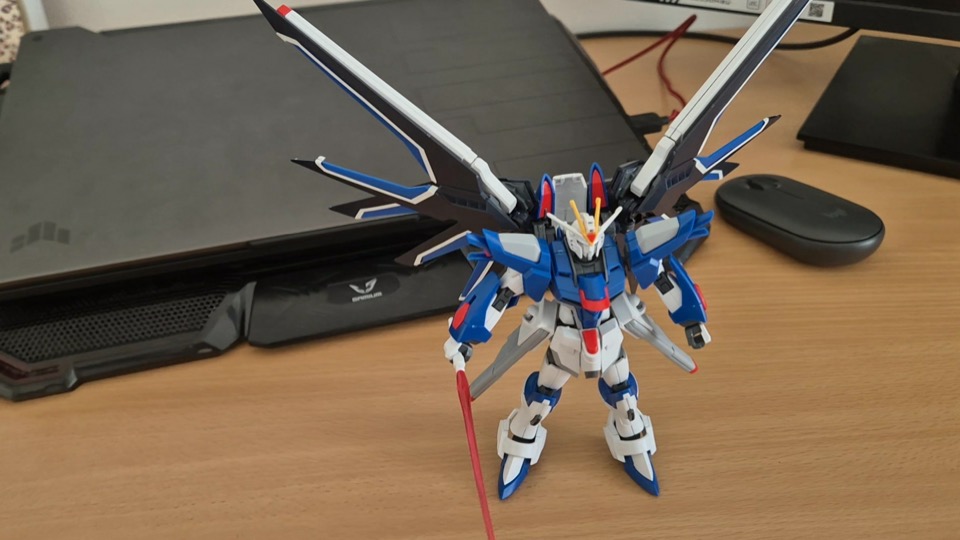}    & \qualimg{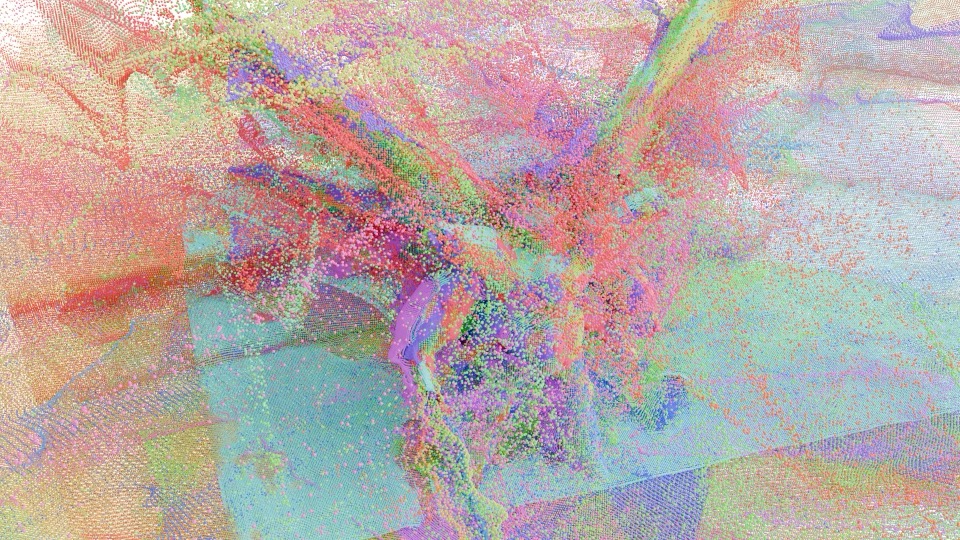}  & \qualimg{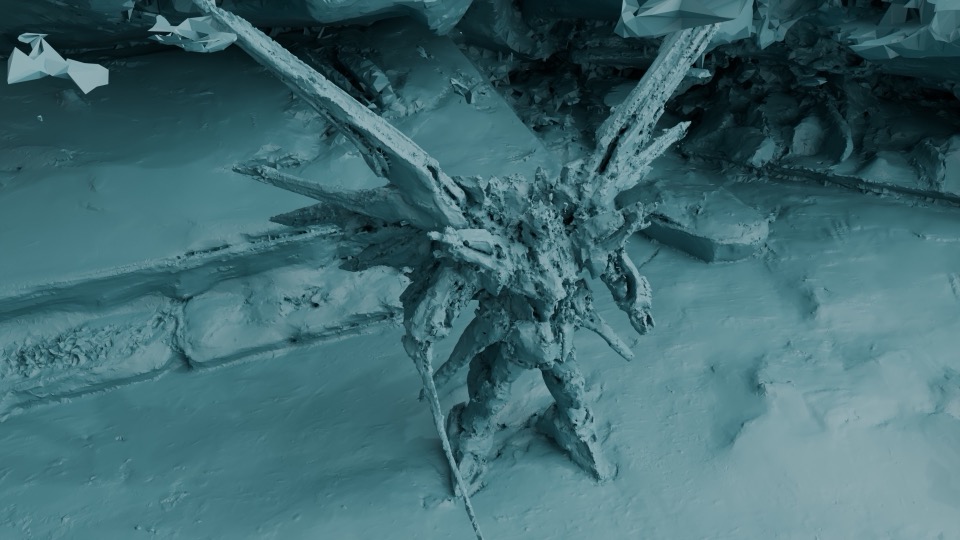}  &
    \qualimg{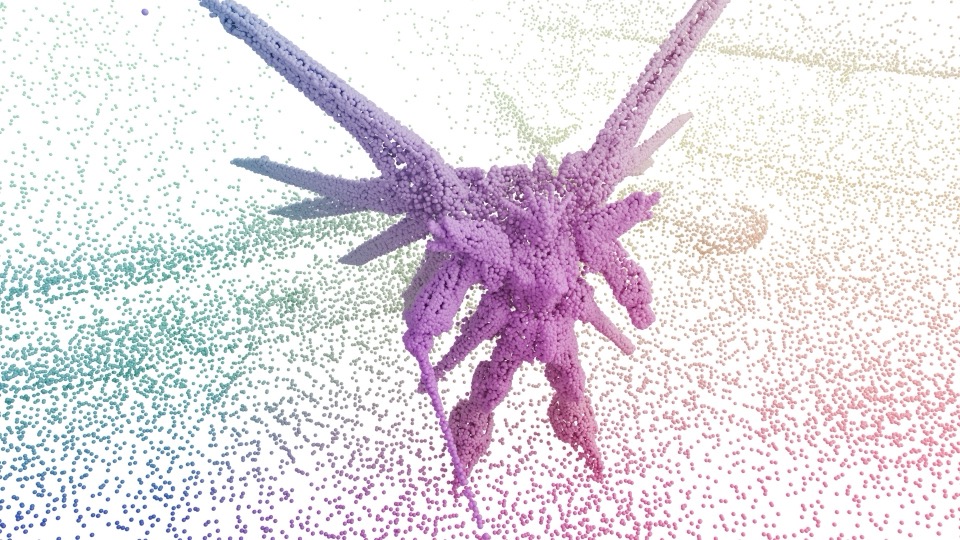} & \qualimg{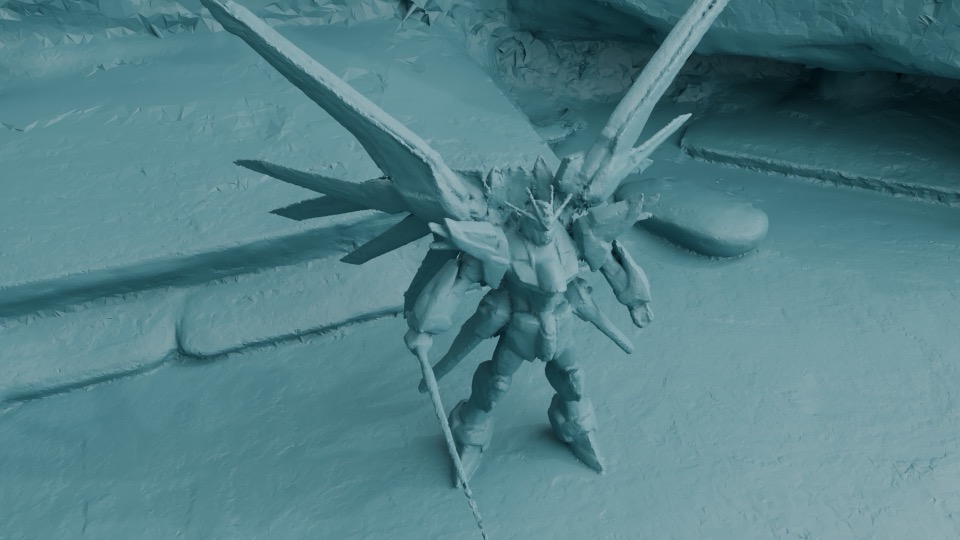} \\
    \qualimg{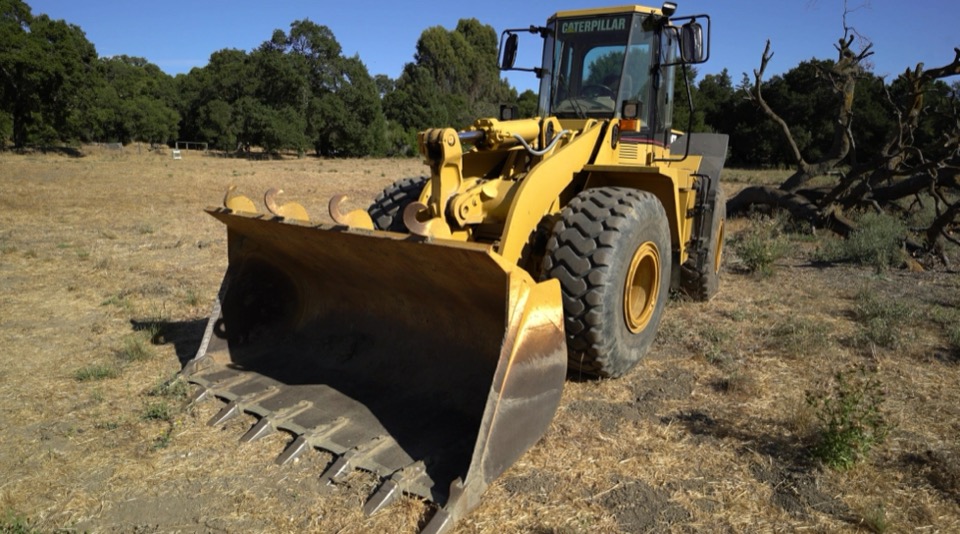}    & \qualimg{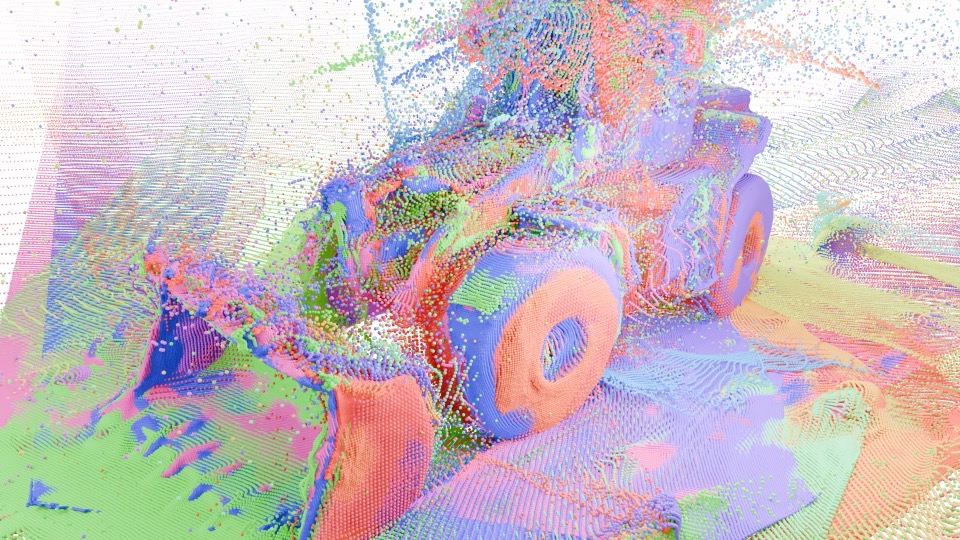}  & \qualimg{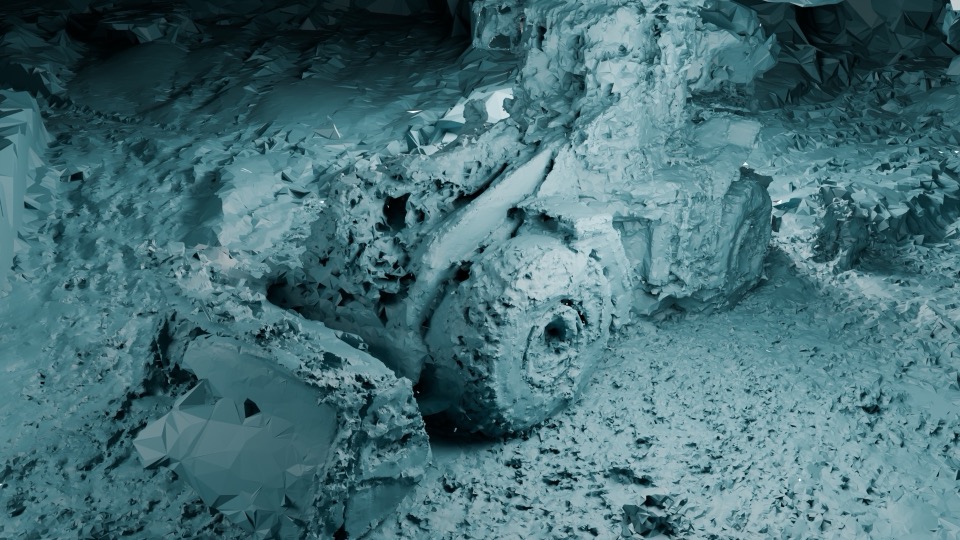}  &
    \qualimg{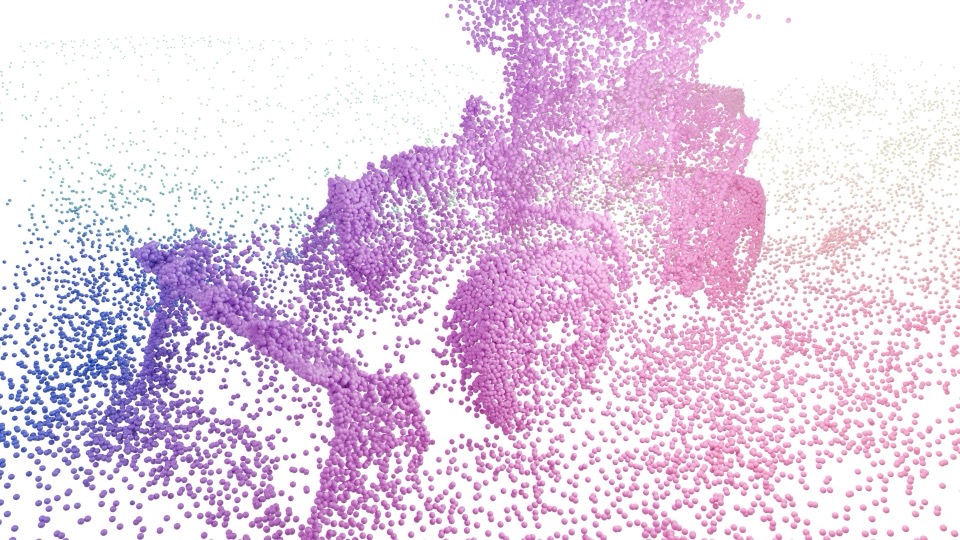} & \qualimg{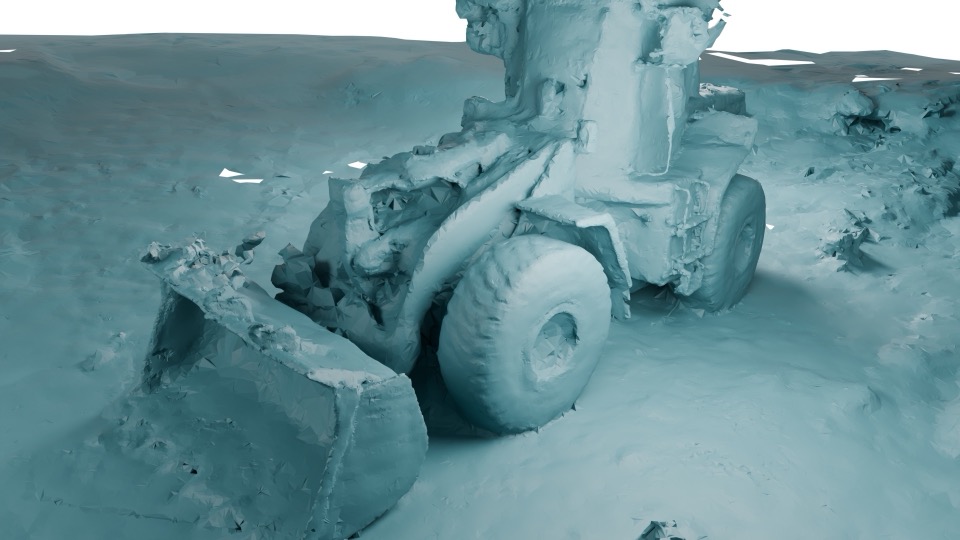} \\
    \qualimg{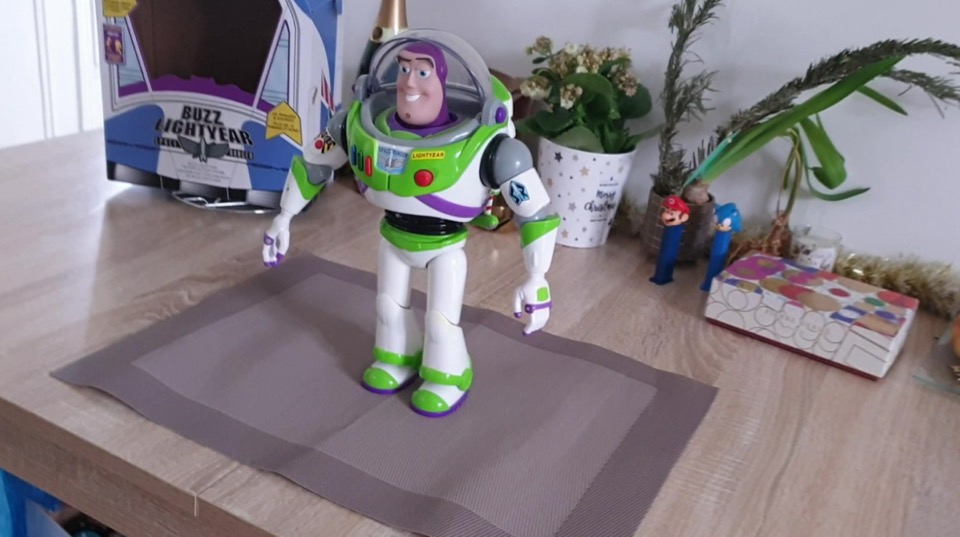}    & \qualimg{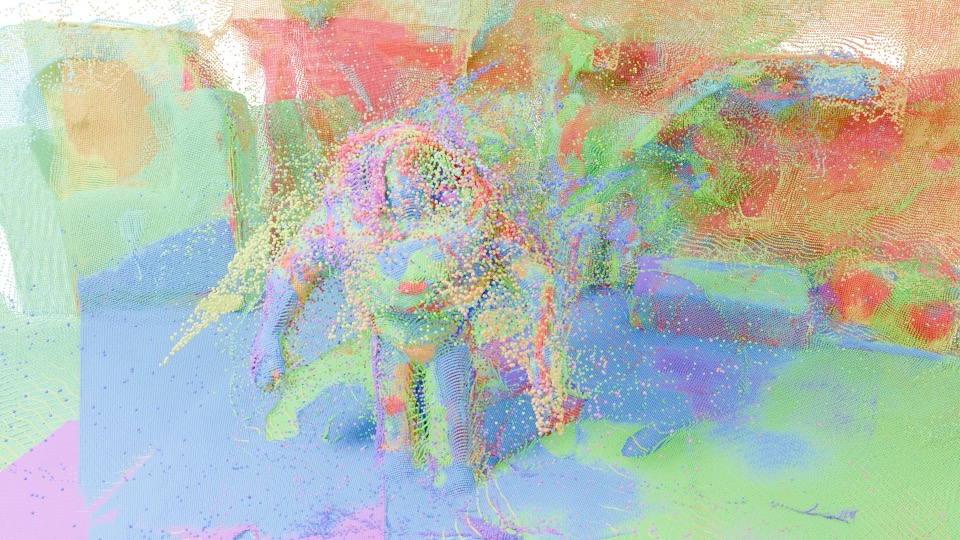}  & \qualimg{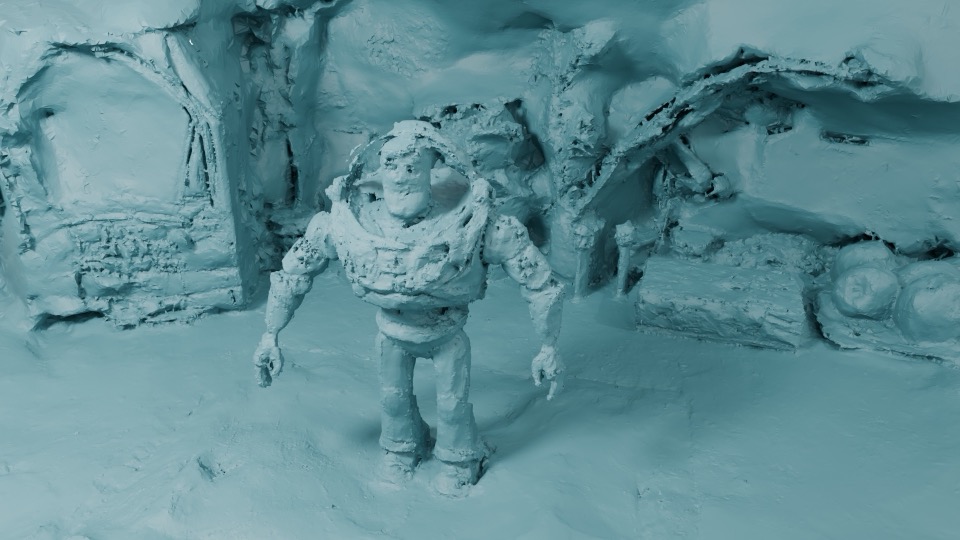}  &
    \qualimg{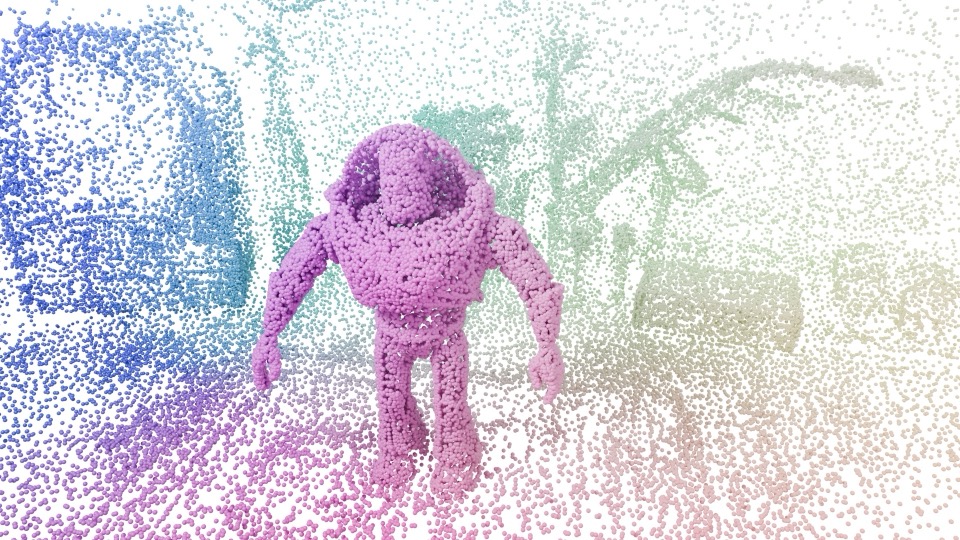} & \qualimg{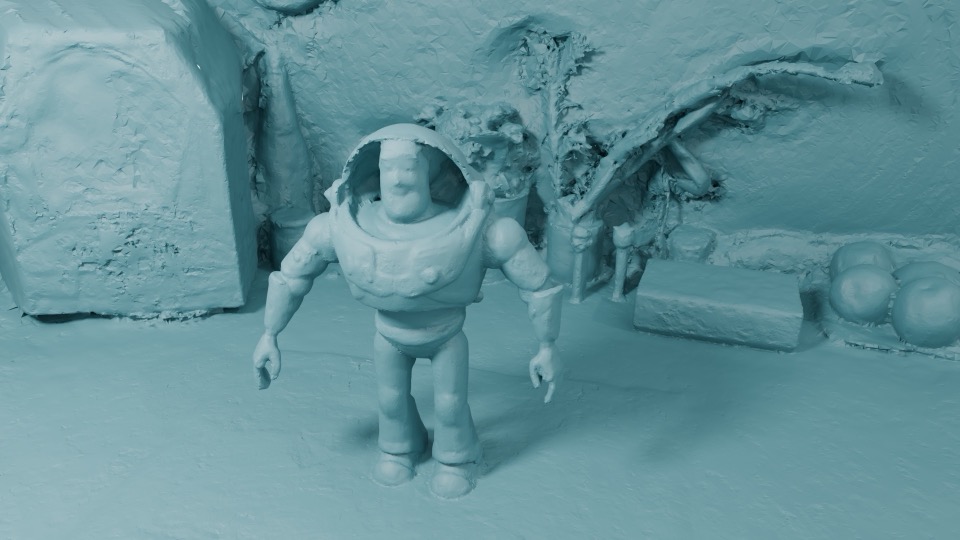} \\
    \qualimg{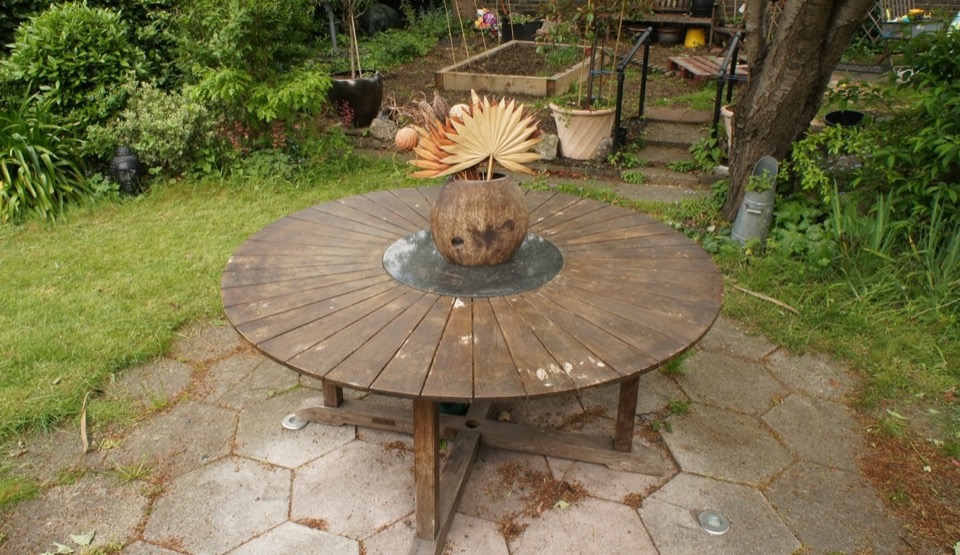}    & \qualimg{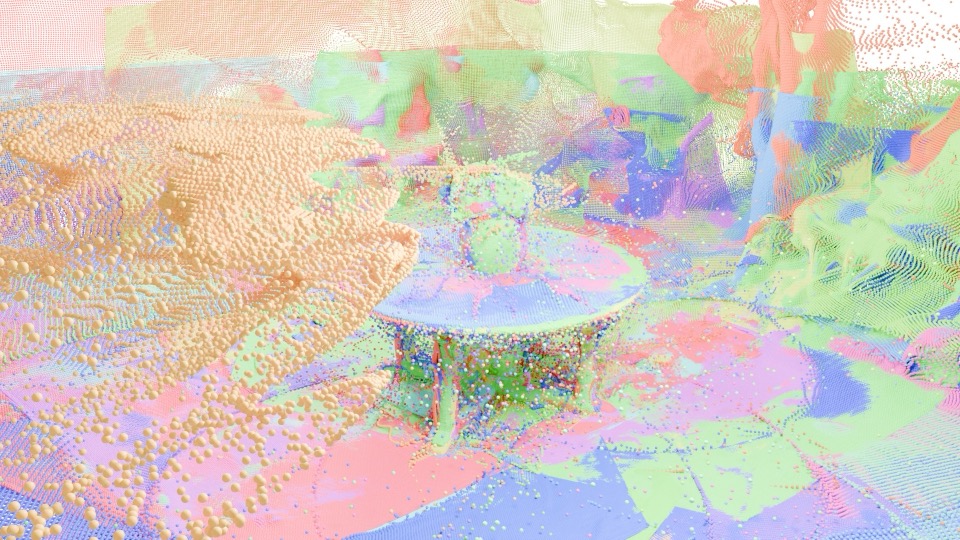}  & \qualimg{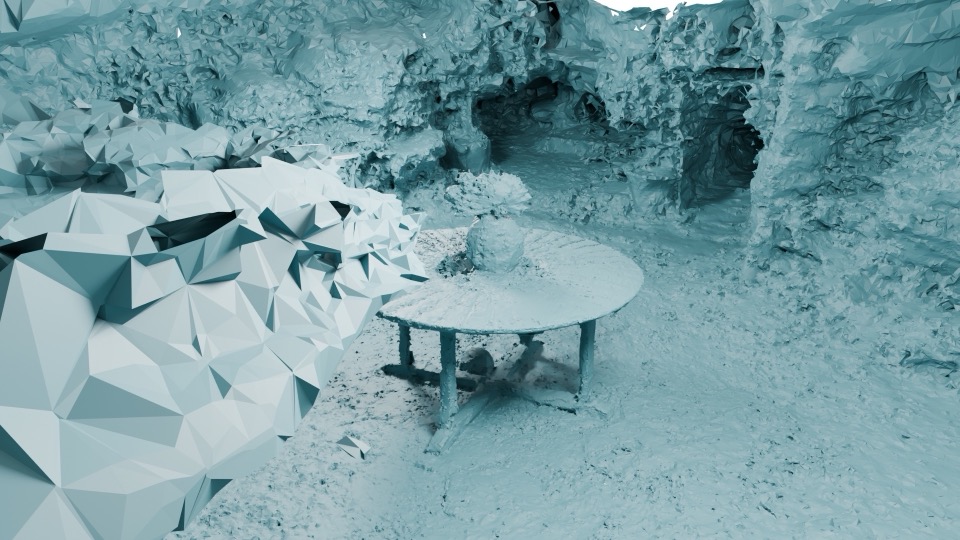}  &
    \qualimg{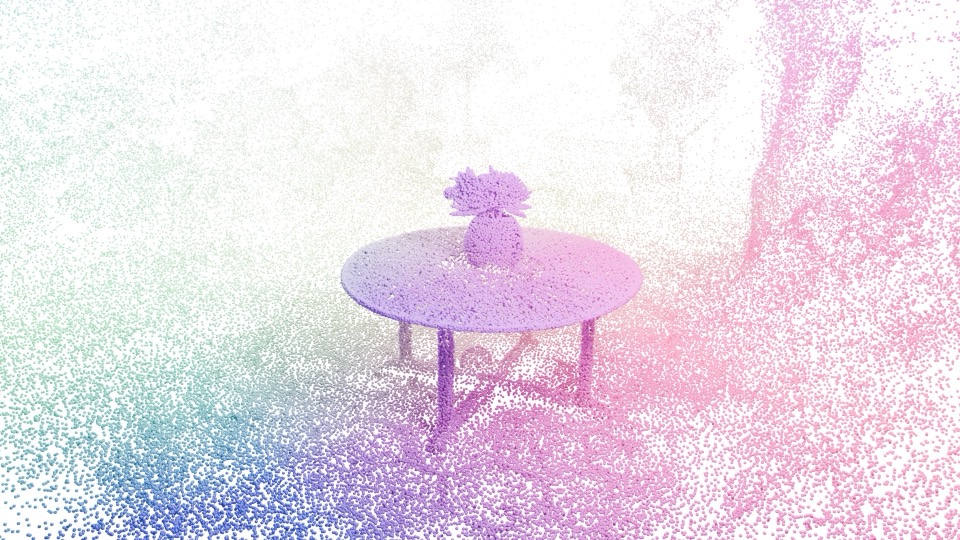} & \qualimg{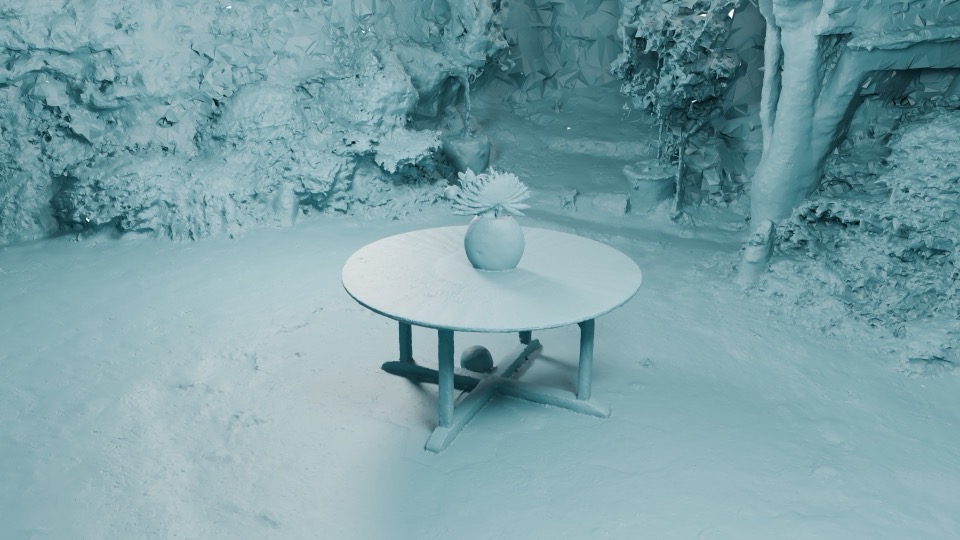} \\
    \qualimg{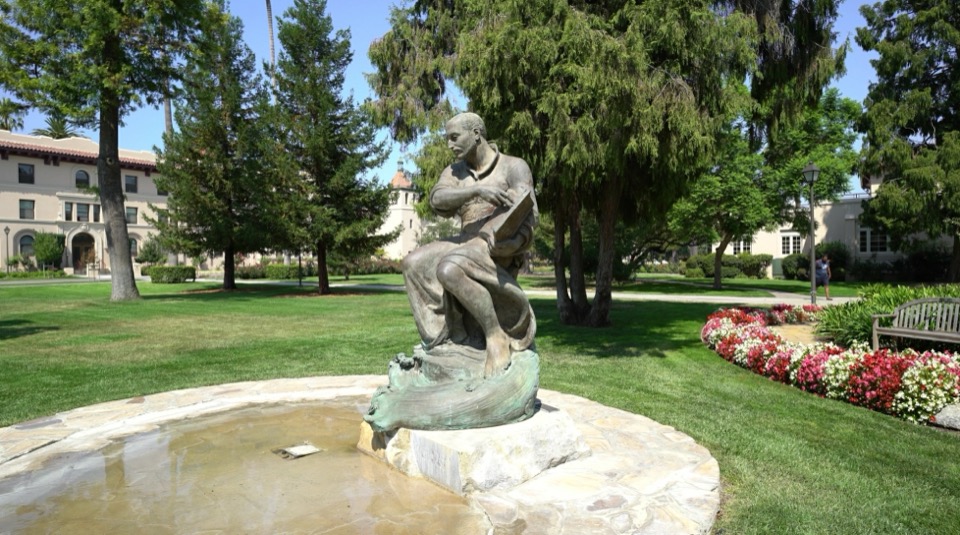}    & \qualimg{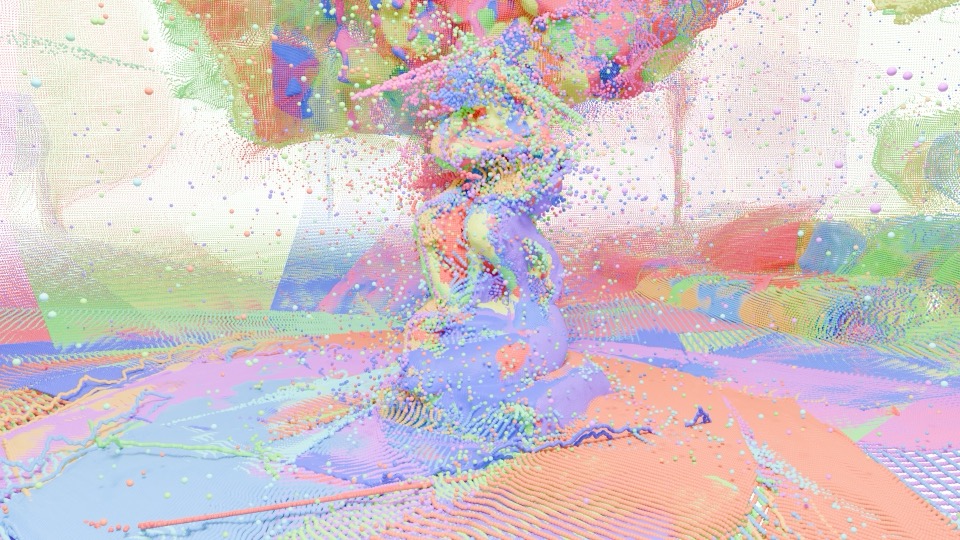}  & \qualimg{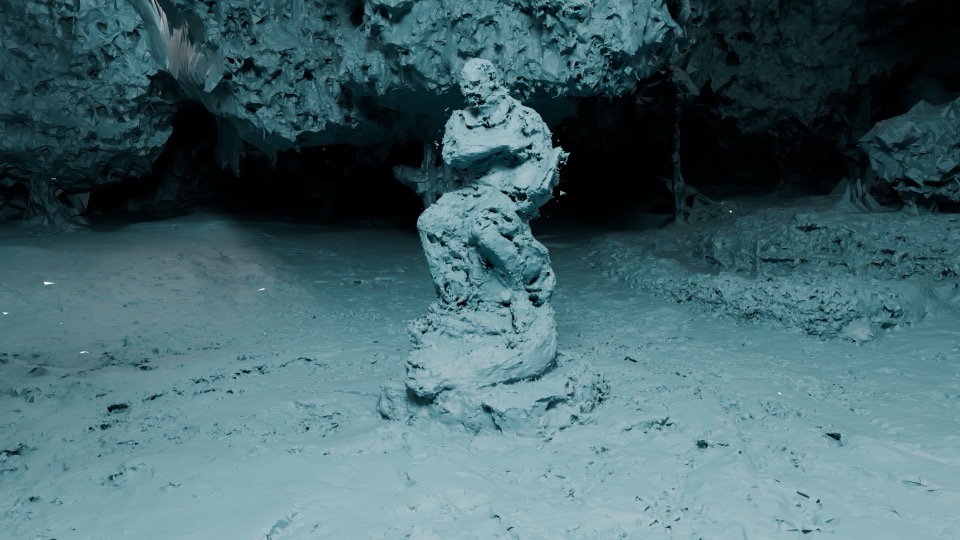}  &
    \qualimg{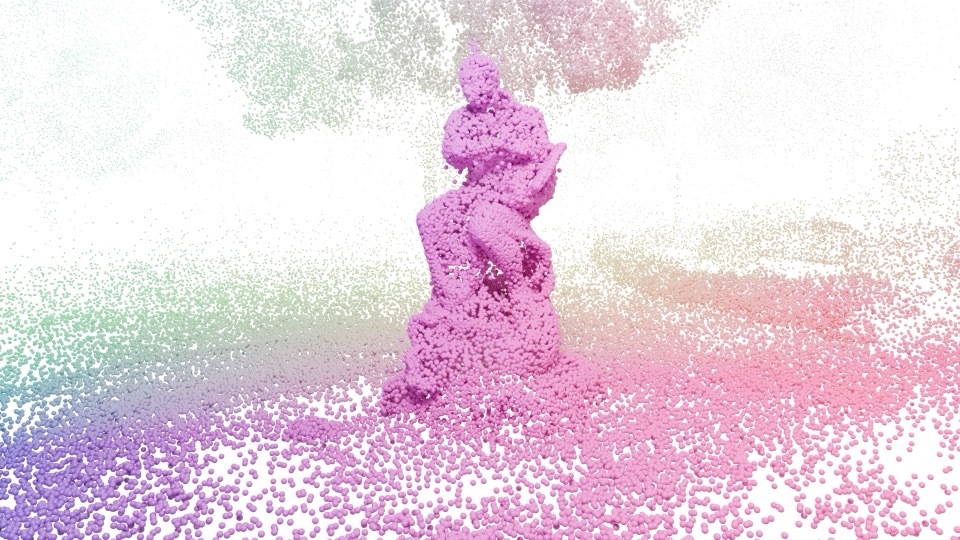} & \qualimg{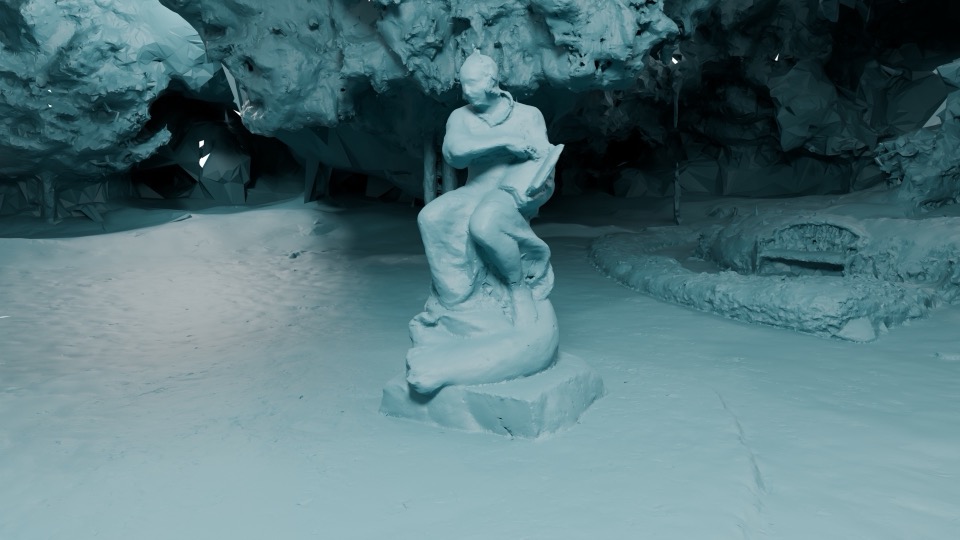} \\
    \qualimg{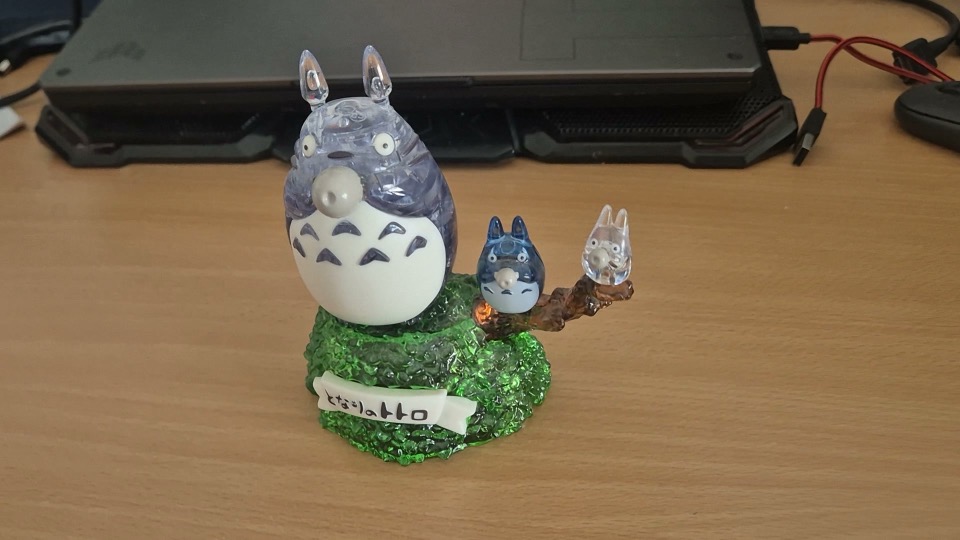}    & \qualimg{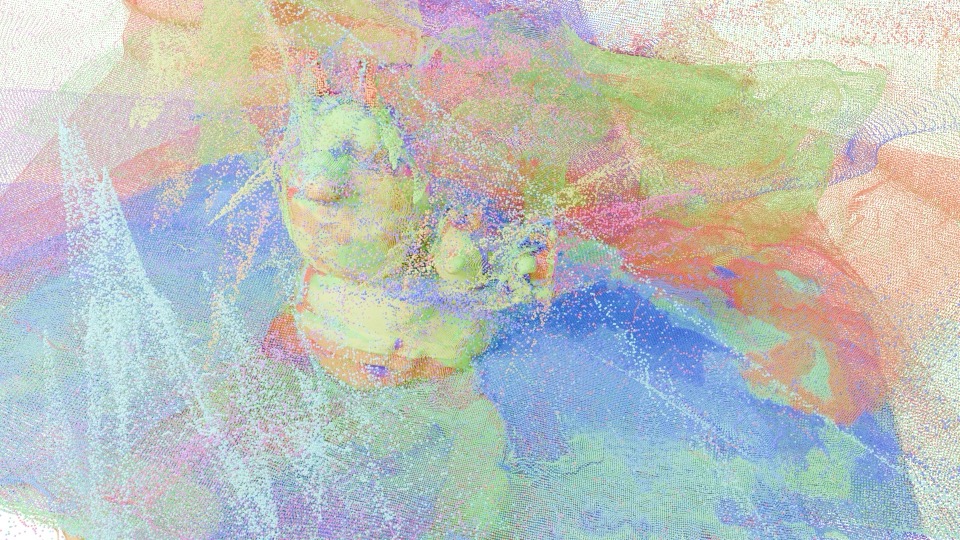}  & \qualimg{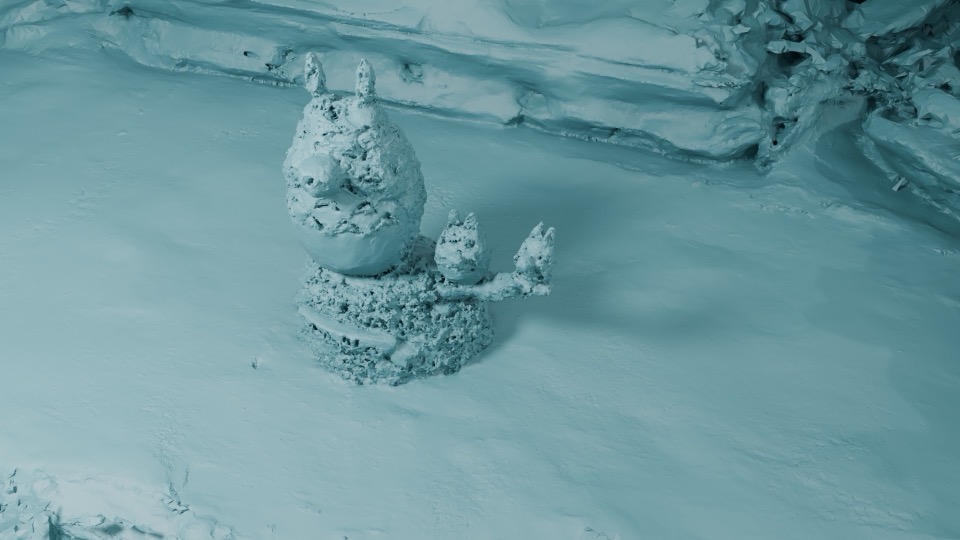}  &
    \qualimg{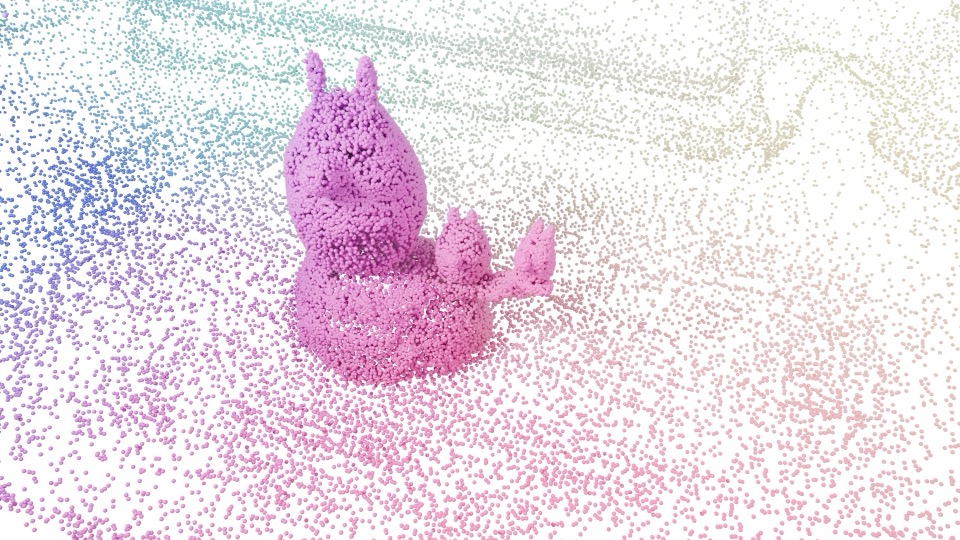} & \qualimg{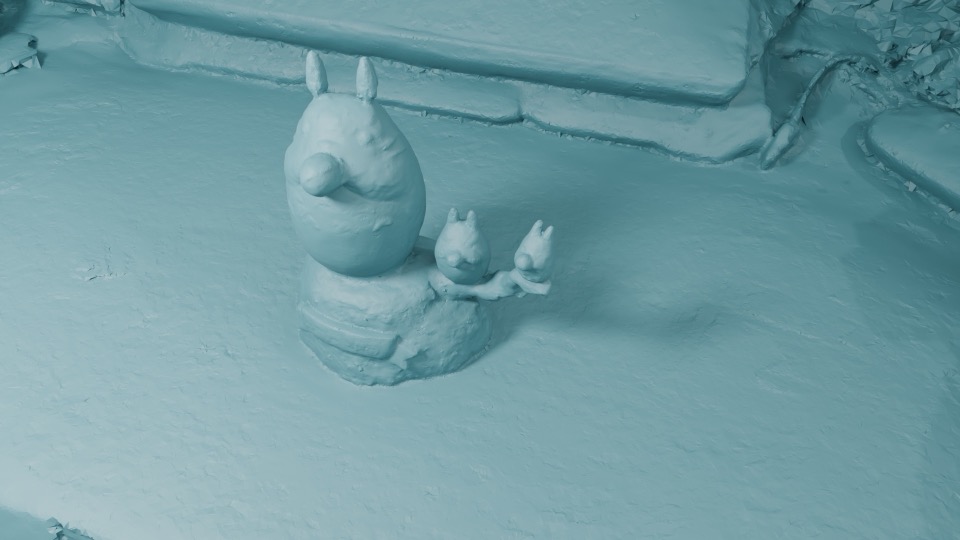} \\
  \end{tabular}
  \caption{\textbf{Qualitative comparisons from $16$ unposed input views.} From left to right: reference view, VGGT~\citep{wang2025vggt} pointmaps (one color per view), Gaussian Wrapping~\citep{gomez2026gaussian} mesh from VGGT initialization, and \methodname's points and mesh. VGGT pointmaps are noisy and duplicated across views, and per-scene GW struggles from few inputs; \methodname yields well-distributed surface points and a clean mesh from the same shared latent.}
  \label{fig:qualitative}
\end{figure}

%% file: tables/pseudo_gt_results.tex
\begin{table}[t]
  \caption{\textbf{Quantitative Evaluation} on four datasets where reference surfaces are generated from Gaussian Wrapping~\citep{gomez2026gaussian} on dense views, covering both foreground and background. DL3DV~\citep{ling2024dl3dv} corresponds to our held-out test split (in-distribution), Tanks~\&~Temples~\citep{knapitsch2017tanks}, Mip-NeRF~360~\citep{barron2022mip} and DeepBlending~\citep{hedman2018deep} are out-of-distribution benchmarks.
  All methods are evaluated from the same set of $16$ unposed input views per scene. Per-view feed-forward baselines come in two flavours: (i) the raw \emph{pointmap} variants (greyed), which simply concatenate per-view depth predictions into a multi-layered point set without enforcing inter-view consistency, and (ii) their TSDF~\citep{curless1996volumetric}-fused counterparts, which collapse the per-view depths into a single global mesh.
  }
  \label{tab:pseudo_gt_results}
  \centering
  \resizebox{\linewidth}{!}{%
  \footnotesize
  \setlength{\tabcolsep}{8pt}
  \begin{tabular}{lcccccccc}
    \toprule
                                          & \multicolumn{2}{c}{DL3DV} & \multicolumn{2}{c}{Tanks\&Temples} & \multicolumn{2}{c}{Mip-NeRF~360} & \multicolumn{2}{c}{DeepBlending} \\
    \cmidrule(lr){2-3}\cmidrule(lr){4-5}\cmidrule(lr){6-7}\cmidrule(lr){8-9}
    Method                                & CD$\downarrow$ & F1$\uparrow$ & CD$\downarrow$ & F1$\uparrow$ & CD$\downarrow$ & F1$\uparrow$ & CD$\downarrow$ & F1$\uparrow$ \\
    \hline
    \multicolumn{9}{l}{\grey \emph{Per-view feed-forward (no global surface)}}                                                                                              \\
    \grey VGGT pointmap~\citep{wang2025vggt}                & \grey 0.0100 & \grey 74.84 & \grey 0.0076 & \grey 81.04 & \grey 0.0132 & \grey 69.18 & \grey 0.0110 & \grey 76.89 \\
    \grey DA3 pointmap~\citep{lin2025depth}                 & \grey 0.0097 & \grey 77.42 & \grey 0.0156 & \grey 71.96 & \grey 0.0144 & \grey 67.59 & \grey 0.0132 & \grey 67.84 \\
    \hline
    \multicolumn{9}{l}{\emph{Per-view feed-forward (fused global surface)}}                                                                                              \\
    VGGT + TSDF~\citep{wang2025vggt}                  & 0.0126 & 69.23 & \tbest 0.0113 & \tbest 77.46 & \tbest 0.0178 & \tbest 60.64 & 0.0193 & 62.30 \\
    DA3 + TSDF~\citep{lin2025depth}                   & \tbest 0.0120 & \tbest 72.30 & 0.0177 & 70.80 & 0.0182 & 59.91 & 0.0210 & 54.03 \\
    \hline
    \multicolumn{9}{l}{\emph{Latent feed-forward (fixed-size output)}}                                                                                              \\
    NOVA3R~\citep{chen2026nova3r}                     & 0.0459 & 30.51 & 0.0432 & 32.99 & 0.0429 & 25.60 & 0.0550 & 27.61 \\
    \hline
    \multicolumn{9}{l}{\emph{Per-scene optimization (sparse-view)}}                                                                                                \\
    2DGS~\citep{huang20242d}                          & 0.0163 & 60.10 & 0.0161 & 62.95 & 0.0222 & 51.08 & 0.0204 & 59.54 \\
    RaDe-GS~\citep{zhang2024radegs}    & 0.0166 & 59.48 & 0.0170 & 61.67 & 0.0224 & 50.83 & 0.0202 & 60.04 \\
    Gaussian Wrapping~\citep{gomez2026gaussian}       & 0.0168 & 60.67 & 0.0157 & 64.94 & 0.0201 & 57.86 & \tbest 0.0164 & \tbest 64.54 \\
    \hline
    \methodname (ours) - No guidance                  & \best 0.0072 & \best 81.92 & \best 0.0053 & \best 88.57 & \best 0.0068 & \best 82.00 & \sbest 0.0116 & \sbest 70.96 \\
    \methodname (ours) - With guidance                & \sbest 0.0083 & \sbest 78.55 & \sbest 0.0056 & \sbest 86.40 & \sbest 0.0103 & \sbest 76.57 & \best 0.0109 & \best 75.09 \\
    \hline
  \end{tabular}%
  }
\end{table}

%% file: tables/main_results.tex
\begin{table}[t]
  \caption{\textbf{Quantitative Evaluation} on four out-of-distribution benchmarks with native surface ground truth: ML-Hypersim~\citep{roberts2021hypersim}, BlendedMVS~\citep{yao2020blendedmvs}, DTU~\citep{aanaes2016dtu}, and SCRREAM~\citep{jung2024scrream}. We report Chamfer Distance (CD, $\downarrow$) and F1-score (F1, $\uparrow$). All methods are evaluated from the same set of $16$ unposed input views per scene. Per-view feed-forward baselines come in two flavours: (i) the raw \emph{pointmap} variants (greyed), which simply concatenate per-view depth predictions into a multi-layered point set without enforcing inter-view consistency, and (ii) their TSDF~\citep{curless1996volumetric}-fused counterparts, which collapse the per-view depths into a single global mesh. 
  }
  \label{tab:main_results}
  \centering
  \resizebox{\linewidth}{!}{%
  \footnotesize
  \setlength{\tabcolsep}{8pt}
  \begin{tabular}{lcccccccc}
    \toprule
                                          & \multicolumn{2}{c}{ML-Hypersim} & \multicolumn{2}{c}{BlendedMVS} & \multicolumn{2}{c}{DTU} & \multicolumn{2}{c}{SCRREAM} \\
    \cmidrule(lr){2-3}\cmidrule(lr){4-5}\cmidrule(lr){6-7}\cmidrule(lr){8-9}
    Method                                & CD$\downarrow$ & F1$\uparrow$ & CD$\downarrow$ & F1$\uparrow$ & CD$\downarrow$ & F1$\uparrow$ & CD$\downarrow$ & F1$\uparrow$ \\
    \hline
    \multicolumn{9}{l}{\grey \emph{Per-view feed-forward (no global surface)}}                                                                                              \\
    \grey VGGT pointmap~\citep{wang2025vggt}            & \grey 0.0070 & \grey 91.91 & \grey 0.0105 & \grey 79.98 & \grey 0.0249 & \grey 47.80 & \grey 0.0063 & \grey 83.36 \\
    \grey DA3 pointmap~\citep{lin2025depth}   & \grey 0.0071 & \grey 87.16 & \grey 0.0739 & \grey 66.74 & \grey 0.0714 & \grey 26.35 & \grey 0.0063 & \grey 83.26 \\
    \hline
    \multicolumn{9}{l}{\emph{Per-view feed-forward (fused global surface)}}                                                                                              \\
    VGGT + TSDF~\citep{wang2025vggt}              & \tbest 0.0138 & \tbest 74.08 & 0.0270 & \tbest 59.64 & 0.0380 & 28.93 & 0.0226 & 59.89 \\
    DA3 + TSDF~\citep{lin2025depth}     & 0.0151 & 69.07 & 0.0875 & 52.61 & 0.0801 & 17.93 & 0.0220 & 60.95 \\
    \hline
    \multicolumn{9}{l}{\emph{Latent feed-forward (fixed-size output)}}                                                                                              \\
    NOVA3R~\citep{chen2026nova3r}                 & 0.0635 & 27.65 & 0.0413 & 32.13 & \tbest 0.0307 & \tbest 31.41 & 0.0771 & 27.41 \\
    \hline
    \multicolumn{9}{l}{\emph{Per-scene optimization (sparse-view)}}                                                                                                \\
    2DGS~\citep{huang20242d}                          & 0.0176 & 62.03 & 0.0295 & 48.19 & 0.0394 & 28.78 & 0.0234 & 53.57 \\
    RaDe-GS~\citep{zhang2024radegs}          & 0.0174 & 62.68 & 0.0303 & 48.85 & 0.0393 & 28.30 & 0.0242 & 53.52 \\
    Gaussian Wrapping~\citep{gomez2026gaussian}  & 0.0145 & 66.86 & \tbest 0.0259 & 55.64 & 0.0460 & 30.17 & \tbest 0.0123 & \sbest 62.96 \\
    \hline
    \methodname (ours) - No guidance                   & \sbest 0.0097 & \sbest 77.98 & \best 0.0103 & \sbest 76.50 & \sbest 0.0242 & \sbest 39.23 & \sbest 0.0114 & \tbest 61.20 \\
    \methodname (ours) - With guidance                   & \best 0.0079 & \best 87.97 & \sbest 0.0114 & \best 77.28 & \best 0.0240 & \best 42.05 & \best 0.0070 & \best 81.11 \\
    \hline
  \end{tabular}%
  }
\end{table}

%% file: figures/multires.tex
\begin{figure}[t]
  \centering
  \newlength{\multiw}\setlength{\multiw}{0.30\linewidth}
  \newlength{\multih}\setlength{\multih}{0.7\multiw}
  \newcommand{\multiimg}[2][\multiw]{%
    \tikz{\node[inner sep=0pt, outer sep=0pt, rounded corners=3pt, clip]
      {\includegraphics[width=#1]{#2}};}%
  }
  \newcommand{\multiph}[2][\multiw]{%
    \setlength{\fboxsep}{0pt}%
    \fcolorbox{imgBorder}{imgFill}{%
      \begin{minipage}[c][\multih][c]%
        {\dimexpr#1-2\fboxrule\relax}\centering
        \footnotesize\itshape\color{black!50} #2
      \end{minipage}%
    }%
  }
  \setlength{\tabcolsep}{2pt}
  \renewcommand{\arraystretch}{1.15}
  \begin{tabular}{@{}c@{\hspace{6pt}}ccc@{}}
                                         &
    \footnotesize $8\mathrm{K}$ points  &
    \footnotesize $32\mathrm{K}$ points &
    \footnotesize $128\mathrm{K}$ points    \\[1pt]
    \rotatebox{90}{\footnotesize \textbf{Points}} &
      \includegraphics[width=\multiw]{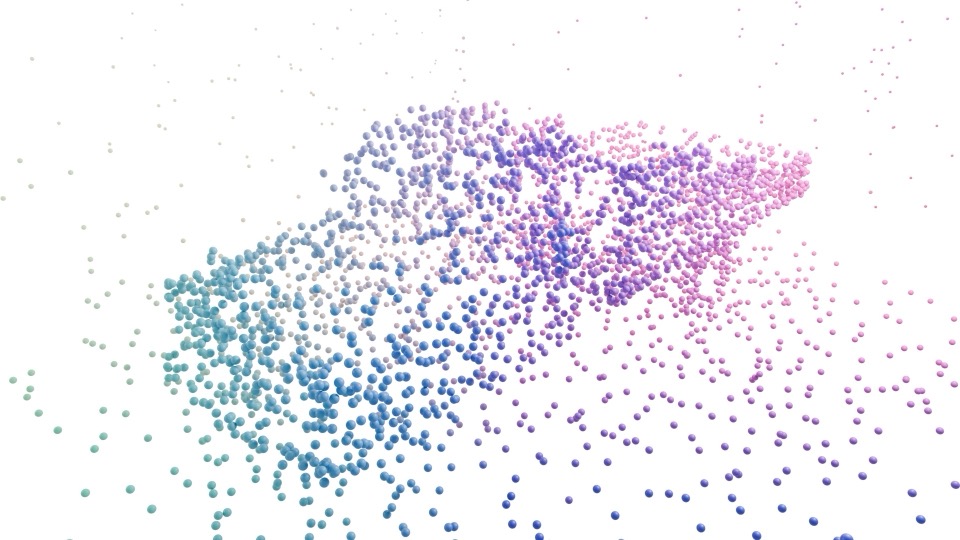}   &
      \includegraphics[width=\multiw]{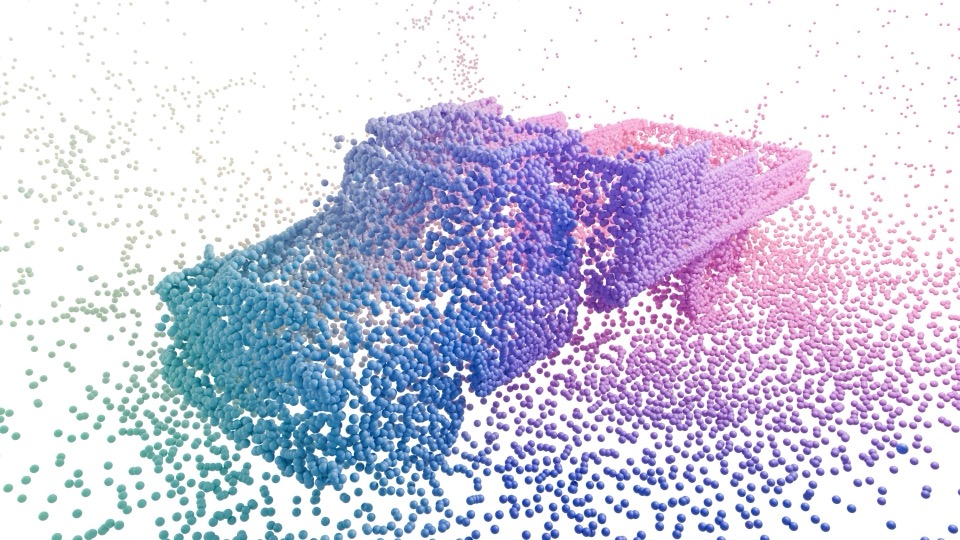}  &
      \includegraphics[width=\multiw]{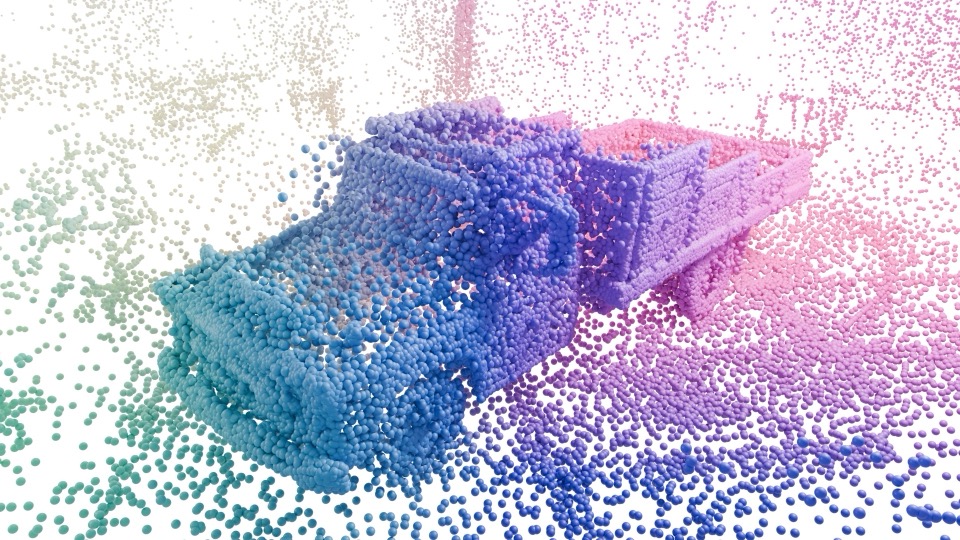} \\
    \rotatebox{90}{\footnotesize \textbf{Mesh}}   &
      \includegraphics[width=\multiw]{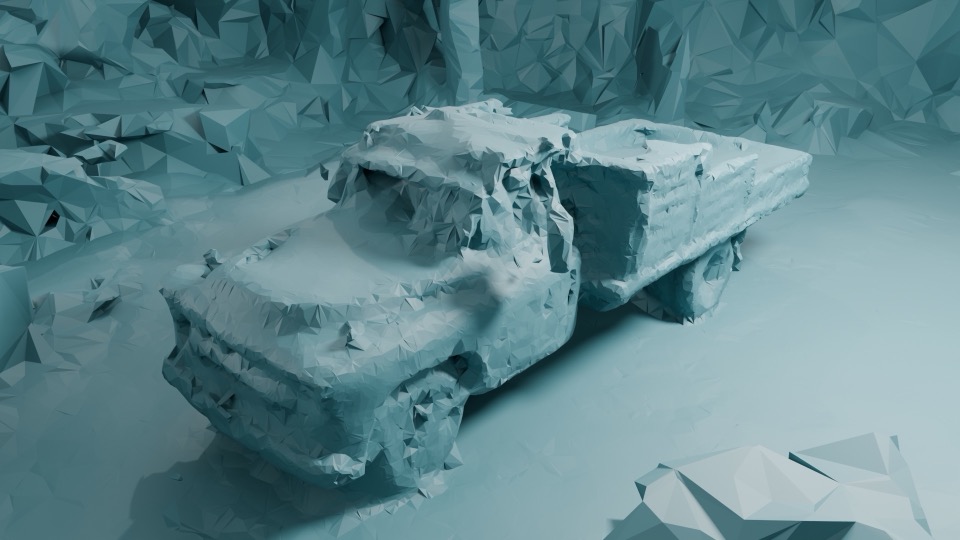}   &
      \includegraphics[width=\multiw]{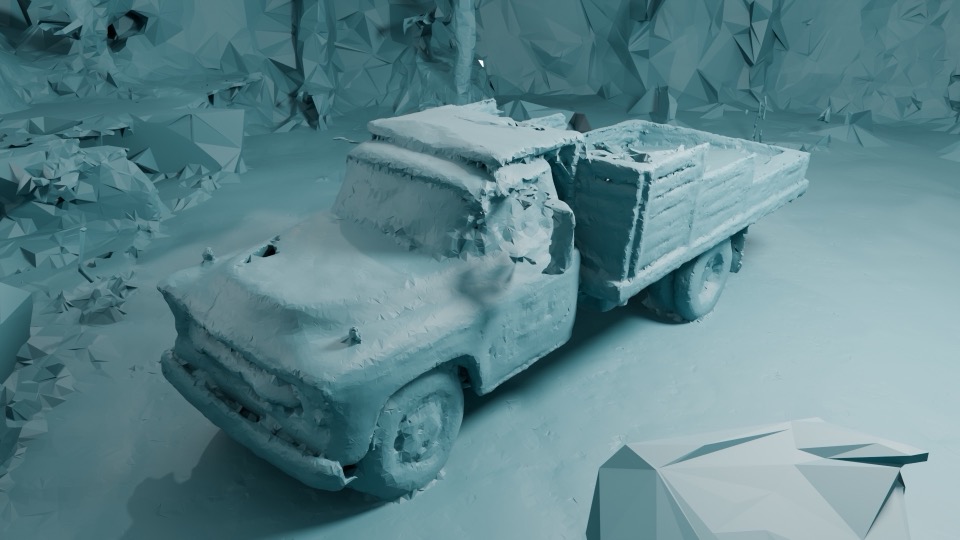}  &
      \includegraphics[width=\multiw]{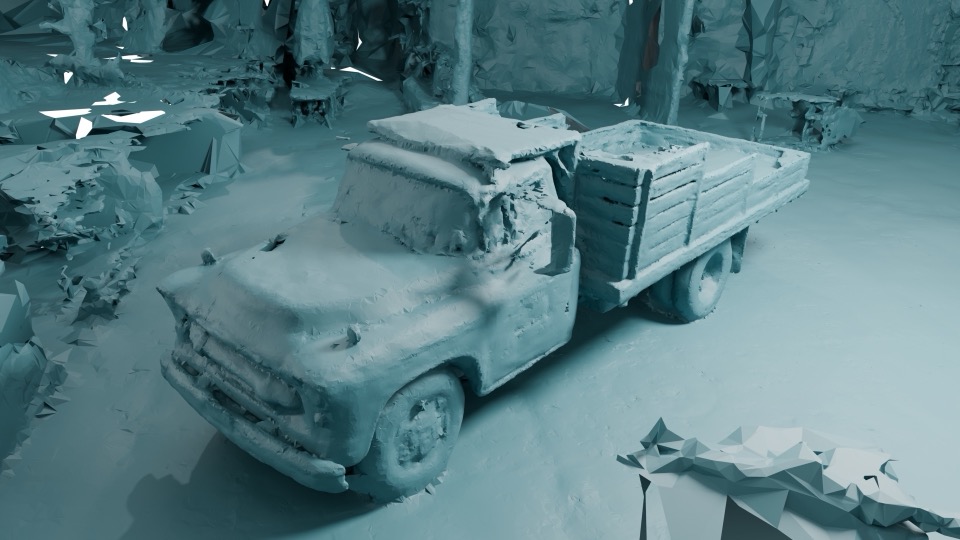} \\
  \end{tabular}
  \caption{\textbf{Variable-resolution decoding from a single shared latent.} Left to right: the flow-matching decoder is queried more densely; \emph{the latent is unchanged across columns}. Bottom: mesh extracted from each point set.}
  \label{fig:multires}
  \vspace{-1em}
\end{figure}

%% file: tables/multiview_tnt.tex
\begin{table}[t]
    \caption{\textbf{Varying number of input views} on Tanks\&Temples. We report Chamfer Distance (CD, $\downarrow$) and F1-score (F1, $\uparrow$) as the number of unposed input views per scene varies from 2 to 32. Best, second-best and third-best are highlighted; greyed per-view pointmap variants are reported for reference and excluded from the ranking.}
    \label{tab:views_tnt}
    \centering
    \resizebox{\linewidth}{!}{%
    \footnotesize
    \setlength{\tabcolsep}{8pt}
    \begin{tabular}{lcccccccc}
      \toprule
                                            & \multicolumn{2}{c}{2 views} & \multicolumn{2}{c}{4 views} & \multicolumn{2}{c}{8 views} & \multicolumn{2}{c}{32 views} \\
      \cmidrule(lr){2-3}\cmidrule(lr){4-5}\cmidrule(lr){6-7}\cmidrule(lr){8-9}
      Method                                & CD$\downarrow$ & F1$\uparrow$ & CD$\downarrow$ & F1$\uparrow$ & CD$\downarrow$ & F1$\uparrow$ & CD$\downarrow$ & F1$\uparrow$ \\
      \hline
      \multicolumn{9}{l}{\grey \emph{Per-view feed-forward (no global surface)}} \\
      \grey VGGT pointmap~\citep{wang2025vggt}            & \grey 0.1461 & \grey 5.27 & \grey 0.0224 & \grey 65.67 & \grey 0.0082 & \grey 80.61 & \grey 0.0064 & \grey 87.44 \\
      \grey DA3 pointmap~\citep{lin2025depth}             & \grey 0.1595 & \grey 6.79 & \grey 0.0247 & \grey 54.88 & \grey 0.0177 & \grey 67.38 & \grey 0.0128 & \grey 74.63 \\
      \hline
      \multicolumn{9}{l}{\emph{Per-view feed-forward (fused global surface)}} \\
      VGGT + TSDF~\citep{wang2025vggt}                    & 0.1444 & 6.83 & \tbest 0.0285 & \tbest 53.62 & \tbest 0.0138 & \tbest 70.85 & \tbest 0.0094 & \tbest 84.83 \\
      DA3 + TSDF~\citep{lin2025depth}                     & \tbest 0.1428 & \tbest 6.97 & 0.0293 & 47.90 & 0.0210 & 61.64 & 0.0140 & 74.59 \\
      \hline
      \multicolumn{9}{l}{\emph{Latent feed-forward (fixed-size output)}} \\
      NOVA3R~\citep{chen2026nova3r}                       & 0.2620 & 5.78 & 0.0502 & 30.29 & 0.0557 & 31.78 & 0.0423 & 35.54 \\
      \hline
      \multicolumn{9}{l}{\emph{Per-scene optimization (sparse-view)}} \\
      2DGS~\citep{huang20242d}                            & 0.1453 & 6.09 & 0.0316 & 43.05 & 0.0187 & 55.95 & 0.0152 & 68.43 \\
      RaDe-GS~\citep{zhang2024radegs}                     & 0.1454 & 6.30 & 0.0314 & 42.60 & 0.0191 & 55.67 & 0.0156 & 66.24 \\
      Gaussian Wrapping~\citep{gomez2026gaussian}         & 0.1476 & 5.24 & 0.0313 & 45.28 & 0.0176 & 60.04 & 0.0133 & 72.10 \\
      \hline
      \methodname (ours) - No guidance                    & \best 0.1345 & \best 9.28 & \best 0.0135 & \best 75.07 & \best 0.0059 & \best 86.59 & \best 0.0049 & \best 90.76 \\
      \methodname (ours) - With guidance                  & \sbest 0.1416 & \sbest 7.08 & \sbest 0.0198 & \sbest 72.65 & \sbest 0.0061 & \sbest 86.25 & \sbest 0.0049 & \sbest 90.34 \\
      \hline
    \end{tabular}%
    }
\end{table}

%% file: tables/multiview_mipnerf.tex
\begin{table}[t]
    \caption{\textbf{Varying number of input views} on Mip-NeRF~360. We report Chamfer Distance (CD, $\downarrow$) and F1-score (F1, $\uparrow$) as the number of unposed input views per scene varies from 2 to 32. Best, second-best and third-best are highlighted; greyed per-view pointmap variants are reported for reference and excluded from the ranking.}
    \label{tab:views_mipnerf}
    \centering
    \resizebox{\linewidth}{!}{%
    \footnotesize
    \setlength{\tabcolsep}{8pt}
    \begin{tabular}{lcccccccc}
      \toprule
                                            & \multicolumn{2}{c}{2 views} & \multicolumn{2}{c}{4 views} & \multicolumn{2}{c}{8 views} & \multicolumn{2}{c}{32 views} \\
      \cmidrule(lr){2-3}\cmidrule(lr){4-5}\cmidrule(lr){6-7}\cmidrule(lr){8-9}
      Method                                & CD$\downarrow$ & F1$\uparrow$ & CD$\downarrow$ & F1$\uparrow$ & CD$\downarrow$ & F1$\uparrow$ & CD$\downarrow$ & F1$\uparrow$ \\
      \hline
      \multicolumn{9}{l}{\grey \emph{Per-view feed-forward (no global surface)}} \\
      \grey VGGT pointmap~\citep{wang2025vggt}            & \grey 0.0765 & \grey 8.33 & \grey 0.0310 & \grey 47.62 & \grey 0.0166 & \grey 67.93 & \grey 0.0133 & \grey 74.90 \\
      \grey DA3 pointmap~\citep{lin2025depth}             & \grey 0.0750 & \grey 8.35 & \grey 0.0300 & \grey 45.45 & \grey 0.0168 & \grey 66.73 & \grey 0.0139 & \grey 73.72 \\
      \hline
      \multicolumn{9}{l}{\emph{Per-view feed-forward (fused global surface)}} \\
      VGGT + TSDF~\citep{wang2025vggt}                    & 0.0755 & \tbest 9.63 & 0.0326 & 40.79 & 0.0244 & \tbest 55.29 & \tbest 0.0161 & \tbest 65.56 \\
      DA3 + TSDF~\citep{lin2025depth}                     & \tbest 0.0746 & 9.24 & \tbest 0.0322 & 40.31 & \tbest 0.0220 & 55.17 & 0.0162 & 64.75 \\
      \hline
      \multicolumn{9}{l}{\emph{Latent feed-forward (fixed-size output)}} \\
      NOVA3R~\citep{chen2026nova3r}                       & 0.0795 & 8.46 & 0.0642 & 21.41 & 0.0492 & 28.01 & 0.0562 & 18.95 \\
      \hline
      \multicolumn{9}{l}{\emph{Per-scene optimization (sparse-view)}} \\
      2DGS~\citep{huang20242d}                            & 0.0754 & 8.73 & 0.0344 & 37.15 & 0.0283 & 47.45 & 0.0217 & 52.13 \\
      RaDe-GS~\citep{zhang2024radegs}                     & 0.0756 & 8.76 & 0.0352 & 37.11 & 0.0293 & 47.89 & 0.0215 & 53.08 \\
      Gaussian Wrapping~\citep{gomez2026gaussian}         & 0.0766 & 7.28 & 0.0339 & \tbest 41.61 & 0.0251 & 53.97 & 0.0162 & 62.41 \\
      \hline
      \methodname (ours) - No guidance                    & \best 0.0714 & \best 13.07 & \best 0.0192 & \best 58.68 & \best 0.0127 & \best 74.07 & \best 0.0071 & \best 81.24 \\
      \methodname (ours) - With guidance                  & \sbest 0.0736 & \sbest 10.40 & \sbest 0.0263 & \sbest 53.71 & \sbest 0.0145 & \sbest 73.44 & \sbest 0.0137 & \sbest 80.66 \\
      \hline
    \end{tabular}%
    }
\end{table}

%% file: figures/nviews.tex
\begin{figure}[t]
  \centering
  \newlength{\nviewW}\setlength{\nviewW}{0.3\linewidth}
  \newlength{\nviewH}\setlength{\nviewH}{0.7\nviewW}
  \newcommand{\nviewimg}[2][\nviewW]{%
    \tikz{\node[inner sep=0pt, outer sep=0pt, rounded corners=3pt, clip]
      {\includegraphics[width=#1]{#2}};}%
  }
  \newcommand{\nviewph}[2][\nviewW]{%
    \setlength{\fboxsep}{0pt}%
    \fcolorbox{imgBorder}{imgFill}{%
      \begin{minipage}[c][\nviewH][c]%
        {\dimexpr#1-2\fboxrule\relax}\centering
        \footnotesize\itshape\color{black!50} #2
      \end{minipage}%
    }%
  }
  \setlength{\tabcolsep}{2pt}
  \renewcommand{\arraystretch}{1.15}
  \begin{tabular}{@{}ccc@{}}
    \footnotesize\textbf{(a) Reference} &
    \footnotesize (b) $3$ views   &
    \footnotesize (c) $9$ views   \\[1pt]
    \includegraphics[width=\nviewW]{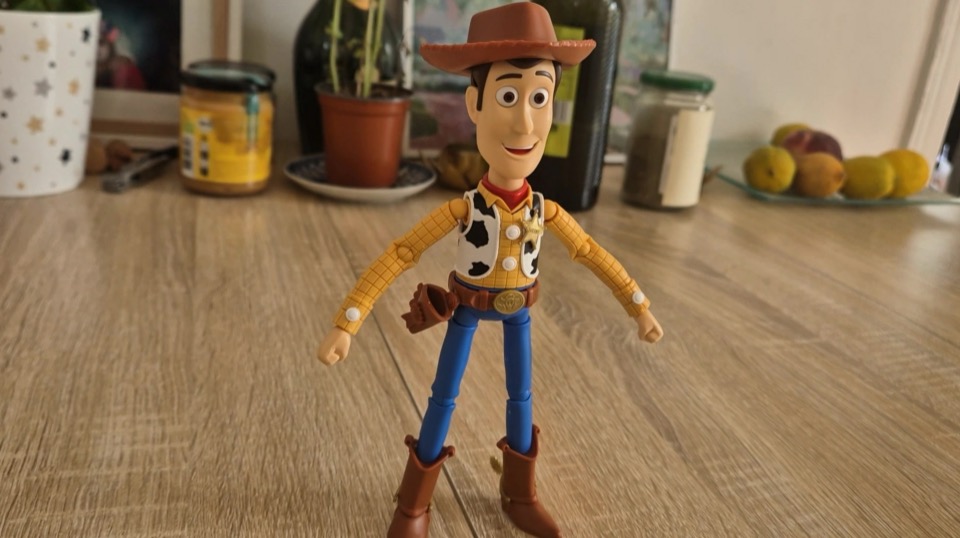} &
    \includegraphics[width=\nviewW]{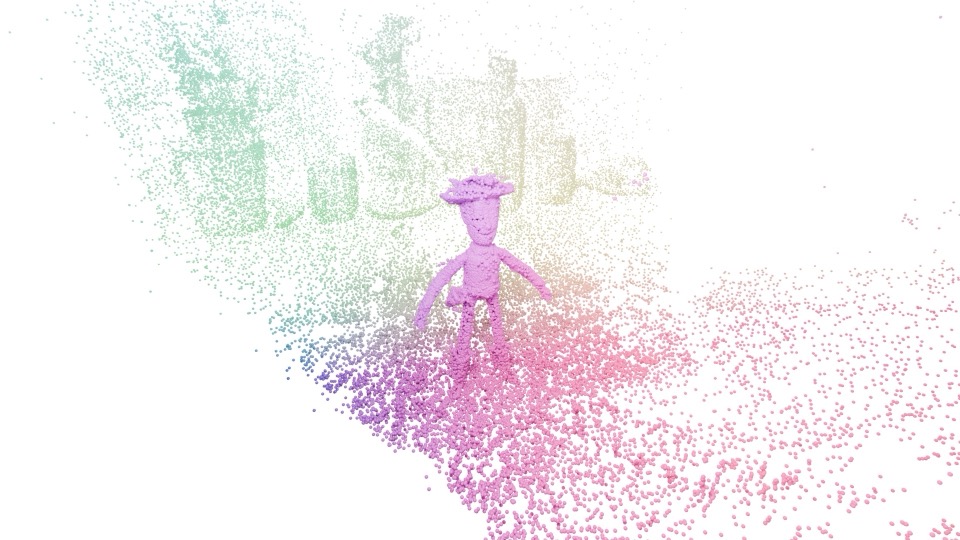}  &
    \includegraphics[width=\nviewW]{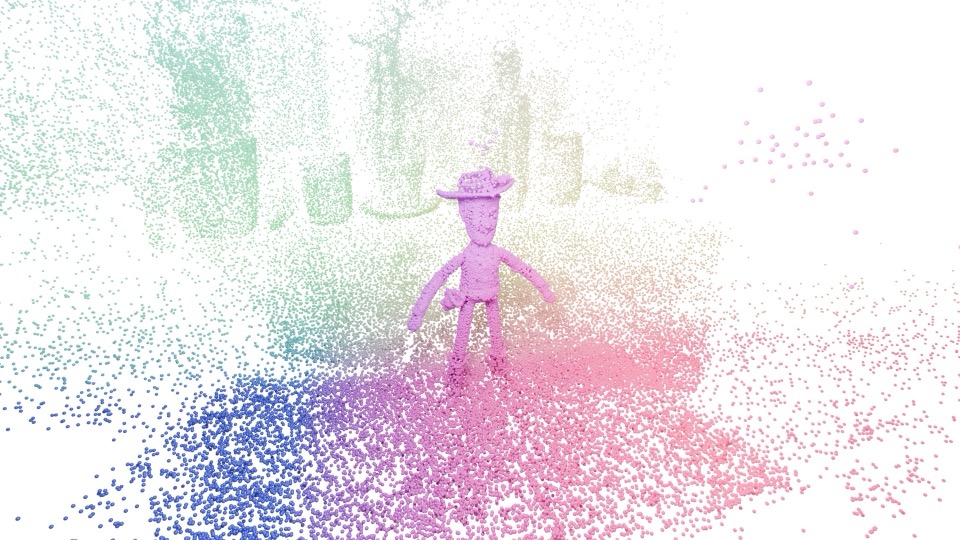}  \\
    \footnotesize (d) $16$ views  &
    \footnotesize (e) $33$ views  &
    \footnotesize (f) $65$ views   \\[1pt]
    \includegraphics[width=\nviewW]{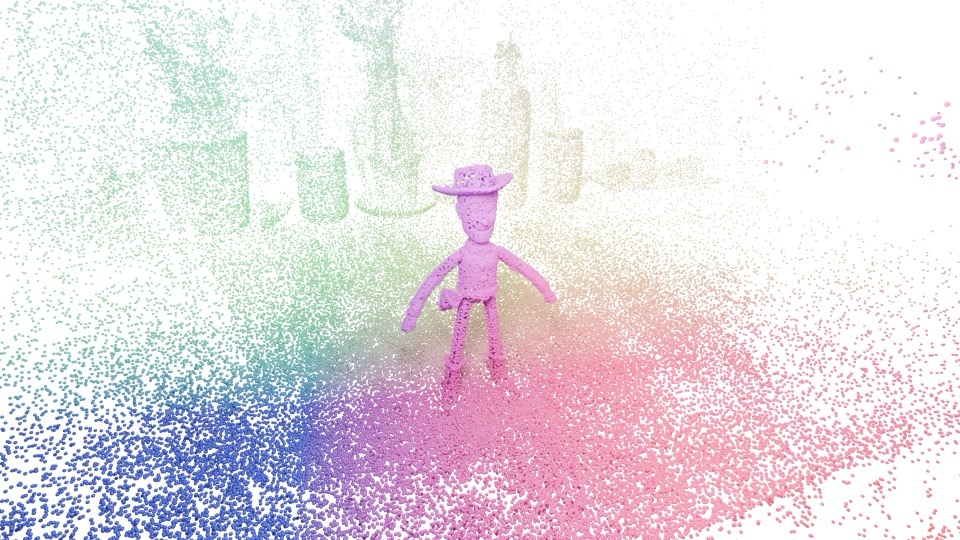} &
    \includegraphics[width=\nviewW]{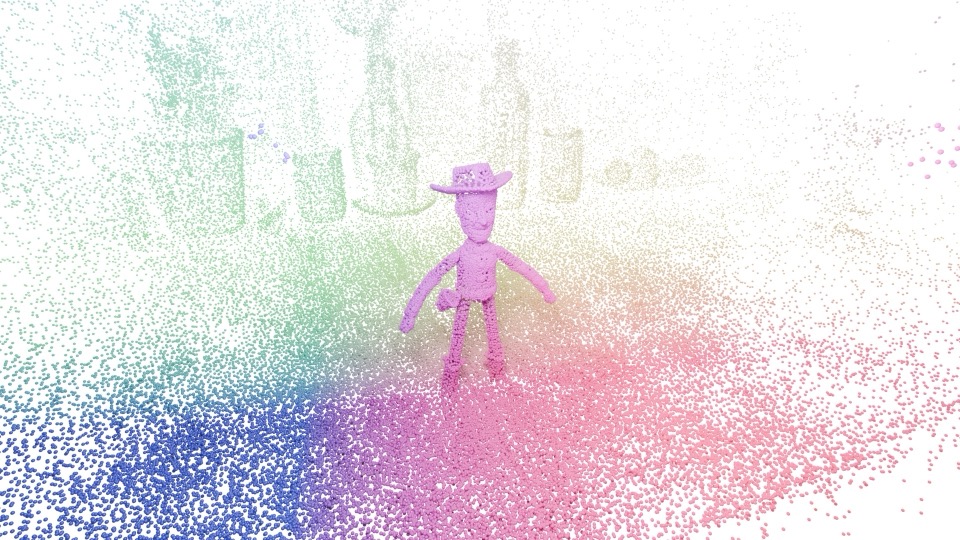} &
    \includegraphics[width=\nviewW]{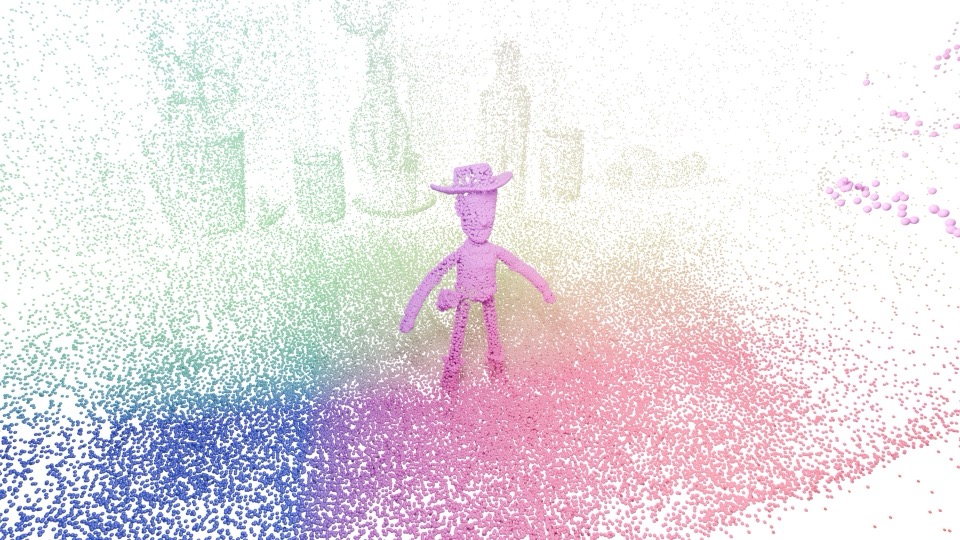} \\
  \end{tabular}
  \caption{\textbf{\methodname{} reconstructions with a varying number of
    input views.} More views progressively complete the scene and sharpen fine detail; the decoder cost is unchanged, since all five reconstructions are produced from a single fixed-size latent. \methodname{} therefore handles a wide range of capture densities without retraining or architectural change.}
  \label{fig:nviews}
\end{figure}

%% file: tables/ablations.tex
\begin{table}[t]
  \caption{Ablation study on the DL3DV test split (in-distribution) and Tanks~\&~Temples (out-of-distribution), with $16$ input views. We vary one design choice at a time and report Chamfer Distance (CD $\downarrow$) and F1 (F1 $\uparrow$).
  }
  \label{tab:ablations}
  \centering
  \resizebox{\linewidth}{!}{%
  \footnotesize
  \setlength{\tabcolsep}{16pt}
  \let\sbestOrig=\sbest
  \def\sbest{}
  \begin{tabular}{llcccc}
    \toprule
                                            &                                           & \multicolumn{2}{c}{DL3DV} & \multicolumn{2}{c}{Tanks~\&~Temples} \\
    \cmidrule(lr){3-4}\cmidrule(lr){5-6}
    Component                               & Variant                                   & CD$\downarrow$ & F1$\uparrow$ & CD$\downarrow$ & F1$\uparrow$ \\
    \hline
    \multirow{3}{*}{Latent size}            & $K = 32$ tokens                           & \sbest 0.0089 & \sbest 72.76 & \sbest 0.0068 & \sbest 79.51 \\
                                            & $K = 128$ tokens                          & \best 0.0073 & \best 81.21 & \best 0.0055 & \best 88.02 \\
    \hline
    \multirow{2}{*}{3D PE on VGGT tokens}   & None (raw VGGT tokens)                    & \sbest 0.0083 & \sbest 76.05 & \sbest 0.0068 & \sbest 80.78 \\
                                            & Gaussian Fourier 3D PE                    & \best 0.0073 & \best 81.21 & \best 0.0055 & \best 88.02 \\
    \hline
    \multirow{2}{*}{Source distribution}    & Pure Gaussian $\mathcal{N}(0,\sigma^2 I)$ & \sbest 0.0088 & \sbest 73.57 & \sbest 0.0074 & \sbest 77.09 \\
                                            & Mixture of Gaussians                      & \best 0.0073 & \best 81.21 & \best 0.0055 & \best 88.02 \\
    \hline
  \end{tabular}%
  \let\sbest=\sbestOrig
  }
\end{table}

%% file: sections/2_related_work_short.tex
\section{Related Work}\label{sec:related}
We provide a focused summary here; a comprehensive discussion is available in Appendix~\ref{sec:related_long}.

\noindent\textbf{Per-view feed-forward 3D.} One class of models predicts geometry—typically depth or pointmaps—separately for each input image. DUSt3R~\cite{wang2024dust3r} and VGGT~\cite{wang2025vggt} produce highly accurate local geometry, but their output is \emph{view-bound}: the number of predicted tokens grows linearly with the number of inputs. These per-view outputs are often redundant and misaligned, making them difficult to fuse into a consistent global mesh~\cite{allshire2025videomimic}. \methodname differs by producing a single, fixed-size latent that represents the entire scene regardless of the number of input views.

\noindent\textbf{Global latent models.} Other works seek a view-agnostic representation. CUT3R~\cite{wang2025cut3r} maintains a persistent state but still decodes it into per-view pointmaps. NOVA3R~\cite{chen2026nova3r} is the closest to our work, using a flow-matching head to decode points from a latent. However, NOVA3R is restricted to a fixed output of $10$K points and is trained on only two views, limiting its ability to reconstruct full scenes. In contrast, \methodname supports decoding an arbitrary number of oriented surface points from a global state.

\noindent\textbf{Flow matching and guidance.} Our decoder builds on flow matching~\citep{lipman2023flow,liu2023flow}, which has shown success in generating point clouds and meshes for single objects~\citep{nichol2022pointe,xiang2025trellis,zhang2026geometrydistributions}. Unlike these methods, which often operate on fixed grids or structured latents, \methodname predicts a velocity field that transports query points independently, reminiscent of MAR~\cite{li2024autoregressive}. This independence is regularized by a guidance mechanism reminiscent of loss-conditioned diffusion~\citep{bansal2023universal,song2023lossguided}. While previous guidance methods typically use external classifiers, our guidance signal is derived directly from a differentiable photometric loss to ensure consistency with the input images.

\noindent\textbf{Per-scene optimization.} High-fidelity reconstruction at the scene scale is traditionally dominated by optimization-based methods like NeuS~\citep{wang2021neus} or 3D Gaussian Splatting~\citep{kerbl2023gaussian}. These methods produce excellent surfaces but require hundreds of \textit{posed} images and significant per-scene compute time. \methodname provides a feed-forward alternative: it produces a reusable scene latent in a single pass, decoupling the cost of the final reconstruction from the number of input images.

%% file: sections/5_conclusion.tex
\section{Conclusion}
\label{sec:conclusion}

We presented \methodname, a feed-forward 3D reconstruction model that compresses an arbitrary number of unposed RGB views into a single fixed-size latent, and decodes an arbitrary number of oriented surface points from that latent through flow matching in $\IR^3\times \mathbb{S}^2$. Per-point independence in the decoder makes the model both flexible and scalable: the same shared latent supports a coarse global preview as well as million-point dense surfaces. To compensate for the lack of explicit surface coupling between independent points, we proposed a photometric guidance scheme that injects the gradient of a differentiable rendering loss into the ODE, correlating nearby points and aligning the output with the input images. As an auxiliary contribution, we built and will release a meshed version of DL3DV~\citep{ling2024dl3dv}, including $\sim$10.5K scenes augmented with watertight surfaces and $10^7$ oriented points each. This dataset made it possible to train \methodname at scale, and we hope it will support future scene-level surface learning. On few-view 3D reconstruction benchmarks, \methodname outperforms strong baselines on standard surface metrics while being the only feed-forward model with this combination of fixed-size latent and arbitrary-resolution decoding.

\noindent\textbf{Limitations.}
\methodname inherits VGGT's failure modes: when the backbone provides poor pointmaps (very few views or extreme baselines), the noisy-pointmap source distribution and VGGT patch tokenization can be unreliable, hurting the decoder's ability to recover. Photometric guidance helps but is not free: rendering oriented Gaussians adds inference cost, though inference time remains small on a single GPU for standard resolutions.
Our supervision comes from Gaussian Wrapping~\citep{gomez2026gaussian} surfaces, which are themselves imperfect on transparent and textureless structures; we believe \methodname will improve as denser and cleaner ground-truth pipelines and larger datasets become available. Finally, the current model represents geometry but not appearance; extending the decoder to also predict view-dependent radiance would be a natural next step.

\section{Acknowledgments}
We are grateful to Maks Ovsjanikov, Patrick Pérez, and Vincent Lepetit for insightful discussions. 
Parts of this work were supported by the ERC Advanced Grant “explorer” (No. 101097259) and JST~ASPIRE~JPMJAP2305.
This work was granted access to the HPC resources of IDRIS under the allocation 2025-AD011013387R3 made by GENCI.

%% file: sections/X_appendix.tex
\input{sections/2_related_work}

\section{Architecture details}
\label{app:arch}

We give in this section the exact hyperparameters of the encoder and decoder used in all main experiments.

\let\shu=\relax

\paragraph{Encoder.}
The frozen VGGT-1B~\citep{wang2025vggt} backbone \shu{(24 transformer blocks, patch size $14$, input resolution $518\times 280$)} produces \shu{$37\times 20 = 740$} patch tokens per view per layer \shu{and $1$ camera token per view,} \shu{all} of dimension $d_{v}=\shu{2048}$. We extract tokens from layers $\ell\in\{4, 11, 17, 23\}$. We then project to a working dimension $D=512$ with a single linear layer\shu{; a separate linear layer projects the camera tokens}.
The 3D Gaussian Fourier features~\citep{tancik2020fourier} use $F=512$ frequencies sampled from $\mathcal{N}(0,\sigma^2 \bI_3)$ \shu{distributed over $16$ log-spaced bands with $\sigma\in\{100, 64.94, \dots, 0.237, 0.154\}$ (geometric ratio $\approx 0.65$), $32$ frequencies per band}, expressed in scene-normalized coordinates \shu{(coordinates are normalized by the median distance to the median VGGT point). The same encoder configuration is shared with the query-point spatial encoder of the decoder, and the resulting 3D PE is broadcast across the four intermediate VGGT layers}. The Perceiver encoder uses $K=128$ learnable latent tokens and $L_\mathrm{s}=\shu{4}$ self-attention blocks with $\shu{16}$ heads \shu{after each of the $L_\mathrm{e}=4$ cross-attention with patch tokens}\shu{, with $\mathrm{mlp\_ratio}=4.0$, qk-norm, LayerScale initialized at $0.01$}.
\shu{The Perceiver encoder for camera tokens applies cross-attention of $1$ latent token with camera tokens for 4 times}\shu{, with $16$ heads and no self-attention. The patch and camera latents are kept on separate paths and consumed by the decoder through different mechanisms (cross-attention for the patch latents, AdaLN conditioning for the camera latent)}.

\paragraph{Decoder.}
The flow-matching decoder is a $L=12$-layer transformer with hidden dimension $D=512$ and $\shu{16}$ heads\shu{, $\mathrm{mlp\_ratio}=4.0$, and qk-norm. There is no self-attention over query points: each block processes points independently, and all spatial information sharing is delegated to the encoder latents}. Each \shu{of the first $6$} block\shu{s} alternates cross-attention to the latent $\bz$ \shu{and an MLP, each of which is prepended with Ada-LN with conditioning tokens (time and camera token). } \shu{The latter $6$ blocks apply only Ada-LN and MLP. Each AdaLN conditioning network produces shift, scale and gate parameters in the DiT~\citep{peebles2023scalable} style ($\mathrm{LN}(x)\cdot(1+\mathrm{scale}) + \mathrm{shift}$, residual gated by $\mathrm{gate}$), with the projection layer zero-initialized so that every block starts as the identity. All AdaLN modules are fed with $\tau(t) + \mathrm{MLP}_{\text{cam}}(\bz_{\text{cam}})$, where $\bz_{\text{cam}}$ is the encoded camera latent and $\mathrm{MLP}_{\text{cam}}$ is a 2-layer 2048-dimensional MLP}. The time embedding $\tau(t)$ uses the standard sinusoidal schedule with $\shu{512}$ frequencies \shu{log-spaced in $[0.1, 1000]$ (output dimension $1024$, projected to $D=512$)}. \shu{Query-point coordinates are normalized per-scene with the same statistics as the patch 3D PE and encoded with the $16$-band Gaussian RFF encoder described above before entering the first block.} A small two-layer MLP head on top of the last block predicts the per-point normal in $\mathbb{S}^2$ alongside the velocity\shu{, yielding a $6$-dimensional output; the final LayerNorm and head are executed in FP32 for precision}.
\shu{Following VGGT~\citep{wang2025vggt}, we use GeLU as the activation function, except for Ada-LN and MLP for the camera token or Fourier-encoded token projections, where we use SiLU.}

\paragraph{Query Formulation.}
For numerical stability, normals are represented as \emph{a residual} within a query point $\bx$. That is, given the sampled 3D point $\mathbf{m} \in \IR^3$ and a normal $\bn \in \mathbb{S}^2$, the query point is initialized as $\bx_0 = (\mathbf{m}, \mathbf{m} + \epsilon \bn) \in \IR^6$, where $\epsilon = 10^{-3}$ multiplied by the scene scale. Velocities are predicted in the same space. 
For decoding the normal from this representation at inference time, we simply take the difference between the second and the first half and normalize the resulting vector.

\section{Training}
\label{app:training}

\input{figures/dataset.tex}

\paragraph{Meshed DL3DV dataset.}
We build a scene-level surface dataset by enriching every scene of DL3DV~\citep{ling2024dl3dv} with a watertight mesh and an associated oriented point cloud. For each scene we run Gaussian Wrapping~\citep{gomez2026gaussian} to extract a watertight surface, and finally sample $\sim$$10^7$ points uniformly on this surface together with their normals. We will release the dataset alongside the paper as an auxiliary contribution. We cache the VGGT tokens once per scene and per view set to avoid recomputing them at every epoch. Figure~\ref{fig:dataset} shows examples from the dataset.

\paragraph{Training.}
The decoder is trained as a velocity-prediction flow-matching model with the conditional optimal-transport scheduler of~\citep{lipman2023flow}. The starting state $\bx_0$ for each query point is sampled as the VGGT world-point prediction (estimated depth values back-projected using estimated camera parameters) plus isotropic Gaussian noise of standard deviation $0.1$ (in scene-normalized coordinates). Time samples are drawn from a logit-normal distribution $t = \sigma(z)$ with $z\sim\mathcal{N}(1.0, 1.6^2)$, biasing supervision towards $t\!\to\!1$. With probability $0.1$ the VGGT features are masked to zero during training to enable classifier-free guidance at inference.

\paragraph{Aligning GT to Predicted Points.} The flow-matching objective supervises each query point against the ground-truth surface, but the two live in different coordinate frames: the ground-truth points are given in the original COLMAP frame of the dataset, whereas the predicted points follow the VGGT convention. We must therefore align them before computing the loss. This alignment has to be accurate enough to preserve fine geometric detail in the supervision, yet fast enough to run at every training step.
We estimate an affine transform $(\mathbf{L}, \mathbf{t})$ that maps COLMAP world coordinates into the VGGT frame, i.e. $\mathbf{p}_{\mathrm{vggt}} \approx \mathbf{p}_{\mathrm{colmap}} \mathbf{L} + \mathbf{t}$. The transform is fitted from depth maps of the ground-truth mesh rendered from the COLMAP cameras, which we precompute for each scene and store as part of the dataset.

We first discard unreliable points by keeping only the top $25\%$ of pixels ranked by VGGT depth confidence, then proceed in two steps.
The first step produces a coarse but robust similarity alignment from the \emph{reference camera alone}: we take only its ground-truth depth points, express them in the reference camera's view space through the COLMAP extrinsics, isotropically rescale them by the ratio of the VGGT and COLMAP camera-baseline extents (the mean distance of the camera centers to their centroid), and move them into the VGGT world frame using the corresponding VGGT camera pose. Because the rotation, translation, and scale are derived from camera geometry rather than from matching depth values, this step is insensitive to per-pixel depth noise and yields a reliable initialization. 
The second step refines this initialization into a full affine transform. For each camera, we measure the residual between the coarsely aligned ground-truth depth points and the VGGT depth points, normalize it by a characteristic spatial scale, and turn it into robust per-point weights via a soft-min; correspondences whose weight falls below the per-camera median are treated as outliers and dropped. A single weighted least-squares fit over the surviving depth points from all cameras then yields the final transform.

\section{Guidance}
\label{app:guidance}

\paragraph{Time grid.} We use a bi-phase time grid of $50+100$ steps with a phase switch at $t=0.95$, sampling $P=10^5$ query points by default. 
At each step $t_k \geq 0.95$ the network predicts the velocity, from which we recover an estimate $\hat{\mathbf{x}}_1$ of the clean sample (point positions and, in the $\mathbb{R}^6$ point--normal parameterization, per-point normals). 
\paragraph{Rendering guidance.} The key idea of \emph{rendering guidance} is to nudge this trajectory, in its final phase, so that the implied surface is photometrically and geometrically consistent with the input images.
Guidance is activated only once $t \ge \tau_{\mathrm{g}} = 0.95$. At activation, the current estimate $\hat{\mathbf{x}}_1$ is used to instantiate a set of 3D Gaussians, one per generated point, whose learnable attributes are anisotropic log-scales, rotation quaternions,
opacity, and view-dependent color (spherical harmonics up to degree~$3$).
\paragraph{Inner loop.} At every subsequent ODE step we run an inner loop of $M=32$ gradient descent iterations that differentiably renders these Gaussians (via the RaDe-GS rasterizer~\citep{zhang2024radegs}) into input views and minimizes a composite rendering loss~\citep{guedon2025matcha,gomez2026gaussian}. 
After the inner loop, the Gaussian displacements define a refined target $\hat{\mathbf{x}}_g$, which we convert back into a velocity; the ODE step is then taken with this \emph{guided} velocity, so that the rendering gradients steer the generative trajectory. We additionally backpropagate the rendering gradients to the camera poses to account for potential, small pose errors in the VGGT predictions.

\paragraph{Monocular geometric experts.} An additional, optional guidance term injects priors from an off-the-shelf monocular expert~\citep{lin2025depth}. 
Following~\citep{guedon2025matcha,guedon2025milo}, we use a scale-invariant depth-order loss between the rendered depth and the monocular depth prior. 
Only relative ordering is enforced, preventing the multi-view inconsistency of monocular depth predictions from affecting the flow trajectory.

\paragraph{Filtering outliers.} The rendering guidance mechanism provides a natural criterion to detect outliers in the flow: noisy points and floaters that do not contribute to any input view converge to a low opacity. We exploit this signal to prune low-opacity points on the fly during the ODE integration, following the standard pruning threshold of 3D Gaussian Splatting~\citep{kerbl2023gaussian}.

\section{Evaluation protocol.}
\label{app:protocol}

To ensure a fair comparison, all baselines receive the same set of unposed views for each scene, and latent methods are evaluated from their official checkpoints. 
Per-view, pointmap-based methods~\citep{wang2025vggt, lin2025depth} do not produce a single surface, so we fuse their predictions with TSDF fusion~\citep{curless1996volumetric}. To recover both foreground and background geometry while preserving fine detail in the foreground, we use a multi-resolution voxel grid: three nested grids of $100^3$ voxels each, with side lengths of $1$, $3$ and $10$ times the scene diagonal, and a TSDF truncation margin of $1.25\%$ of the scene diagonal.

\paragraph{Point cloud alignment.} To compare a predicted point cloud against the ground-truth surface, we first bring the prediction into the ground-truth coordinate frame and then measure geometric agreement. All distance thresholds are expressed as fractions of the ground-truth bounding-box diagonal $d = \lVert \mathbf{p}_{\max} - \mathbf{p}_{\min} \rVert$, so that the protocol is invariant to the absolute scale of each scene.
Alignment proceeds in two stages. 
First, we estimate a closed-form $7$-DoF similarity transform (scale, rotation, and translation) with the Umeyama algorithm~\citep{umeyama1991least}, using the predicted and ground-truth camera centers as correspondences; this fixes the global scale. Second, we refine the rigid component with a scale-locked, robust point-to-point ICP, in which the Umeyama scale is held fixed and only rotation and translation are updated. 
ICP runs for up to $30$ iterations on voxel-downsampled clouds (voxel size $10^{-3}d$). 
In each iteration, correspondences beyond a truncation distance are discarded, where the truncation radius follows a geometric schedule from $5\times10^{-2}d$ down to $10^{-2}d$ so that early
iterations are permissive and later ones strict. 
Of the surviving correspondences we retain only the closest $70\%$ (least-trimmed squares), weight their residuals with a Huber loss (with $\delta$ set to half the current truncation radius), and solve a weighted Procrustes problem for the rigid update. 

\paragraph{Metrics.} After alignment, we clip prediction points lying outside the axis-aligned ground-truth bounding box to avoid asymmetrically inflating the prediction-to-ground-truth distance term. Finally, both clouds are voxel-downsampled to a common density (voxel size $10^{-3}d$), and we report the symmetric Chamfer distance together with the F-score at a distance threshold $\tau = 10^{-2}d$. Chamfer distances are reported normalized by $d$ to allow for aggregation across scenes of differing absolute scale.

%% file: sections/2_related_work.tex
\section{Related work}
\label{sec:related_long}

\paragraph{Per-view feed-forward 3D geometry.}
The first family of feed-forward 3D models predicts a pointmap or depth map per input image. DUSt3R~\cite{wang2024dust3r} pioneered this paradigm by regressing paired pointmaps in a common frame from two views, and \citet{leroy2024grounding} extended it to more accurate matching. More recent work scales the idea to many views with a single transformer: VGGT~\cite{wang2025vggt} processes hundreds of images in one forward pass,  \citet{lin2025depth} pushes the formulation further with a depth–ray prediction target,  \citet{yang2025fast3r} parallelises DUSt3R-style decoding across $1000{+}$ views, and \citet{wang2026pi3} removes the dependency on a fixed reference frame.
Streaming variants such as Spann3R~\citep{wang2024spann3r} and ingredients such as metric monocular pointmaps~\citep{wang2025moge,wang2025moge2}, prior-conditioned regression~\citep{jang2025pow3r}, and unified factored representations~\citep{keetha2026mapanything} further enrich this space. 
These models are remarkably accurate per view, but the output remains \emph{view-bound}: the number of predicted tokens grows linearly with the number of inputs, the per-view pointmaps overlap heavily, and they are rarely consistent enough to be fused into a clean mesh~\cite{allshire2025videomimic}.

\paragraph{Latent and global feed-forward 3D.}
A second line of work attempts to keep a single, view-agnostic representation.
CUT3R~\cite{wang2025cut3r} maintains a persistent state through cross-attention but still reads it out with per-view metric pointmaps.
Concurrent global tokenization approaches such as 3DRAE~\citep{wei2025any} target this goal but, in practice, still query the latent through per-view decodings. Recent work NOVA3R~\cite{chen2026nova3r} is the closest to our setting: it compresses unposed multi-view inputs into a fixed-size latent and decodes points with a flow-matching head, but emits a fixed number of 3D points with no ability to sample more or less. Furthermore, it is trained with at most 2 input views, limiting its ability to fully reconstruct the entire scene.
Closely related in spirit, D4RT~\citep{zhang2026efficientlyd4rt} also encodes the input into a single global scene representation that is decoded by independent point queries. It is, however, designed for \emph{dense video}: its encoder operates on a temporally-patchified frame sequence to target 4D reconstruction, and it decodes by regressing the 3D position of \emph{pixel-anchored} spatio-temporal queries rather than generatively transporting free oriented points onto a surface. \methodname instead targets sparse images and decodes arbitrary 3D points through flow matching.
Generalisable Gaussian-splatting models follow a related philosophy: pixelSplat, MVSplat, NoPoSplat~\citep{charatan2024pixelsplat,chen2024mvsplat,ye2025noposplat} and the LRM family~\citep{hong2024lrm,zhang2024gslrm,ziwen2025longlrm,szymanowicz2024splatter} predict per-pixel or per-view 3D Gaussians from a few sparse images, sharing the redundancy issue of pointmap methods.
On the architectural side, our encoder is a Perceiver~\citep{jaegle2021perceiver,jaegle2022perceiver}, an architecture that has been repurposed as a generative backbone in recurrent interface networks (RIN)~\citep{jabri2022scalable, dufour2024don, dufour2025miro, haji2026one} for image generation; we adapt it here to feed-forward 3D by compressing multi-view tokens into a fixed-size latent that is then exposed through a decoder that can sample arbitrary number of oriented points directly in the 3D space.
\methodname differs from this entire family in that it produces a single, fixed-size, view-independent latent \emph{and} exposes the geometry through an arbitrary-resolution decoder.

\paragraph{Per-scene optimization for surfaces.}
At the scene scale, fine-grained 3D reconstruction is still dominated by per-scene optimization. 
One branch targets photo-realistic novel-view synthesis without explicit surfaces: NeRF~\citep{mildenhall2020nerf} and its accelerated or anti-aliased successors~\citep{muller2022instant,barron2022mip,barron2023zip}, 3D Gaussian Splatting~\citep{kerbl2023gaussian}, and recent variants such as Radiant Foam~\citep{govindarajan2025radiant}, all of which fit a radiance field that produces faithful renderings but no clean geometry. 
The other branch is explicitly geometry-focused: NeuS~\citep{wang2021neus} and VolSDF~\citep{yariv2021volume} render an implicit signed-distance field, while SuGaR~\citep{guedon2024sugar}, 2D Gaussian Splatting~\citep{huang20242d}, Gaussian Opacity Fields~\citep{yu2024gaussian}, RaDe-GS~\citep{zhang2024radegs}, MILo~\citep{guedon2025milo} and Gaussian Wrapping~\citep{gomez2026gaussian} build on radiance-field representations to extract a manifold-quality mesh, often combined with classical surface-extraction machinery~\citep{lorensen1987marching,curless1996volumetric,kazhdan2006poisson}.
Both branches reach high fidelity but share the same practical limits: they rely on dense captures of hundreds of images, require tens of minutes of per-scene optimization, and yield no reusable latent. Sparse-view variants such as \citet{guedon2025matcha} relax the input requirement, yet their time and memory cost scales linearly with the number of views, limiting their scalability.
\methodname instead leverages Gaussian Wrapping~\citep{gomez2026gaussian} only at training time, as a source of dense surface supervision, and is itself feed-forward and scene-level at inference: it decodes surfaces at arbitrary resolution from a single shared latent from unposed input images, decoupling the cost of reconstruction from the number of input views.

\paragraph{Object-level learned 3D and shape latents.}
Feed-forward learned mesh predictors offer an alternative to optimization, but remain largely \emph{object}-level: LRM and its Gaussian-splatting variant GS-LRM~\citep{hong2024lrm,zhang2024gslrm} regress NeRFs or 3DGS from one or a few posed views, InstantMesh~\citep{xu2024instantmesh} adds an iso-surface extraction head, and TRELLIS~\citep{xiang2025trellis} learns a structured latent for object generation with a rectified-flow transformer. Sam3D~\cite{chen2025sam} predicts 3D geometry and texture from a single image.
Closer to our own design, 3DShape2VecSet~\citep{zhang20233dshape2vecset}, Michelangelo~\citep{zhao2023michelangelo} and CLAY~\citep{zhang2024clay} represent shapes as a small set of latent vectors  decoded by cross-attention,
and a wide ecosystem of 3D-native generators built on this idea reconstructs single objects from images or text~\citep{liu2023zero123,liu2023one2345,long2024wonder3d,zhao2025hunyuan3d}.
None of these methods targets full scenes from a few unposed views with a shared, queryable latent, which is precisely \methodname's regime.

\paragraph{Diffusion / flow priors and guidance.}
A growing line of work uses 2D diffusion as a prior for novel-view synthesis or 3D reconstruction: GeNVS~\citep{chan2023genvs},  CAT3D~\citep{gao2024cat3d} and ReconFusion~\citep{wu2024reconfusion} hallucinate plausible views that are
then turned into geometry through an external pipeline, NeRFbusters~\citep{warburg2023nerfbusters} works directly in 3D as a voxel-based regulariser, and ReconViaGen~\citep{chang2025reconviagen} distils generative priors into object reconstruction. 
Score-distillation methods~\citep{poole2023dreamfusion,wang2023prolificdreamer} push the same idea further at the price of long per-scene optimization. Their target is pixels or single objects rather than a global, queryable scene representation. The closest mathematical tool to ours is flow matching and its relatives~\citep{lipman2023flow,liu2023flow,albergo2023building,albergo2023stochastic}, which build on diffusion foundations~\citep{ho2020denoising,song2021denoising,song2021scorebased,karras2022elucidating,rombach2022high,peebles2023scalable} and were already shown to be effective for point clouds and meshes in PointFlow, LION, Point-E, {Geometry Distributions}, and TRELLIS~\citep{yang2019pointflow,vahdat2022lion,nichol2022pointe,zhang2026geometrydistributions,xiang2025trellis}.
Instead of denoising images, our decoder learns a velocity field that transports query points in $\IR^3$ \textit{independently} from a noisy source to the surface, conditioned on the scene latent. {Conceptually, this design echoes MAR~\citep{li2024autoregressive} in image generation: MAR pairs a global autoregressive prior over image tokens with a per-token diffusion head, sampling each patch conditioned on a global state; \methodname analogously pairs a global feed-forward prior, built on VGGT~\citep{wang2025vggt}, with a per-point flow-matching head, replacing patch-wise transport in pixel space with point-wise transport in $\IR^3$. 
The guidance mechanism we use during ODE integration is reminiscent of classifier and classifier-free guidance~\citep{dhariwal2021diffusion,ho2021classifier} and of training-free inverse-problem solvers ~\citep{chung2023diffusion,bansal2023universal,song2023lossguided}, but here the gradient signal comes from a differentiable photometric rendering of the partially solved point cloud rather than from an external classifier or likelihood.

%% file: figures/dataset.tex
\begin{figure}[t]
    \centering
    \newlength{\datasetW}\setlength{\datasetW}{0.3\linewidth}
    \newlength{\datasetH}\setlength{\datasetH}{0.7\datasetW}
    \newcommand{\datasetimg}[2][\datasetW]{%
      \tikz{\node[inner sep=0pt, outer sep=0pt, rounded corners=3pt, clip]
        {\includegraphics[width=#1]{#2}};}%
    }
    \newcommand{\datasetph}[2][\datasetW]{%
      \setlength{\fboxsep}{0pt}%
      \fcolorbox{imgBorder}{imgFill}{%
        \begin{minipage}[c][\datasetH][c]%
          {\dimexpr#1-2\fboxrule\relax}\centering
          \footnotesize\itshape\color{black!50} #2
        \end{minipage}%
      }%
    }
    \setlength{\tabcolsep}{2pt}
    \renewcommand{\arraystretch}{1.15}
    \begin{tabular}{@{}ccc@{}}
      &
      \footnotesize (a) Reference &
      \\[1pt]
      \includegraphics[width=\datasetW]{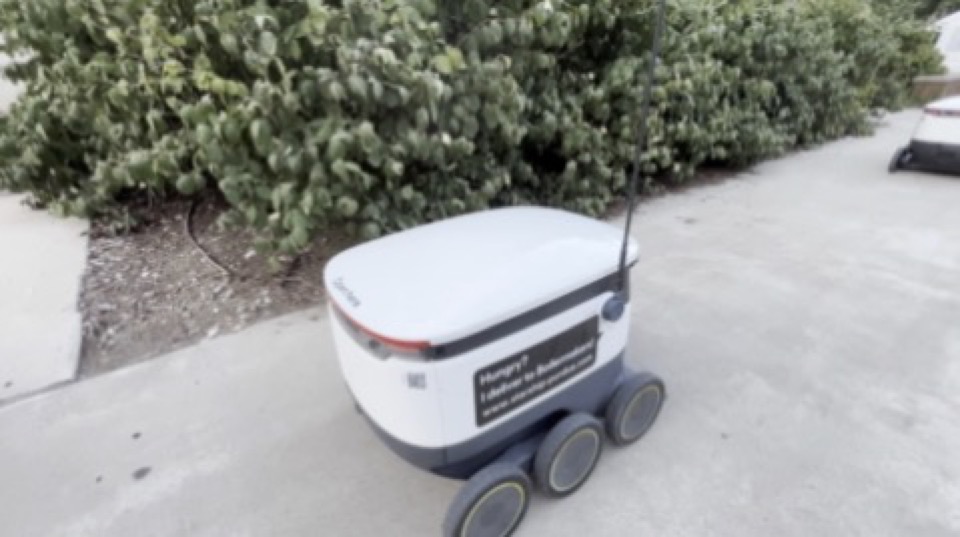} &
      \includegraphics[width=\datasetW]{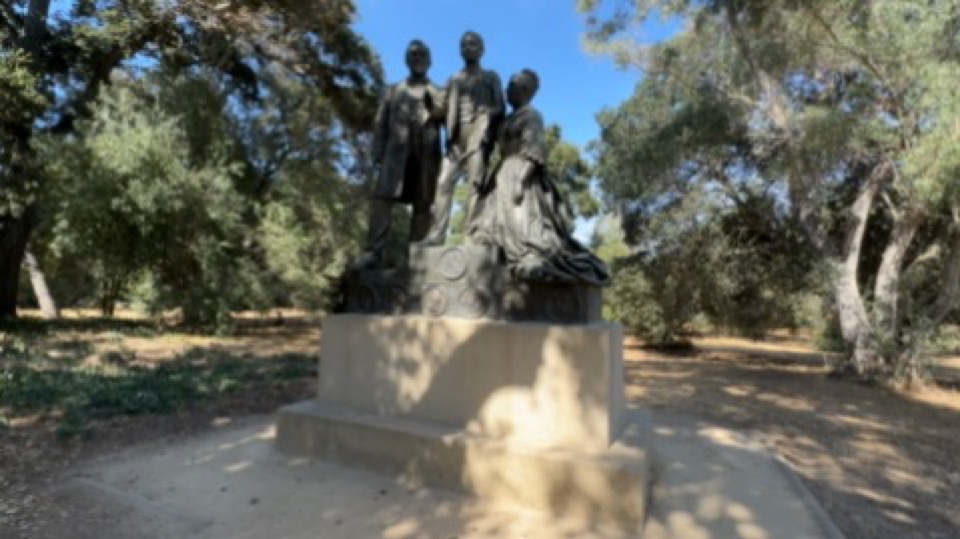}  &
      \includegraphics[width=\datasetW]{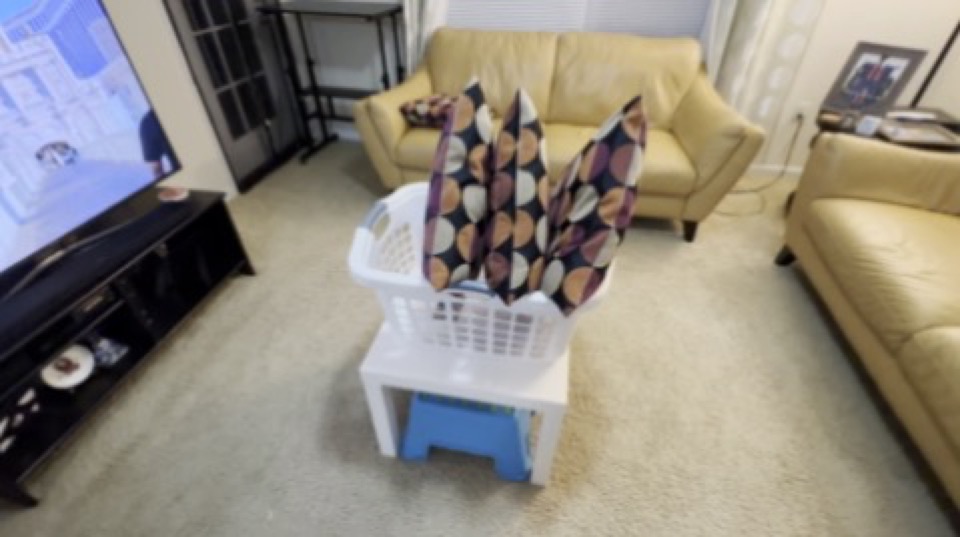}  \\
      &
      \footnotesize (b) Watertight mesh &
      \\[1pt]
      \includegraphics[width=\datasetW]{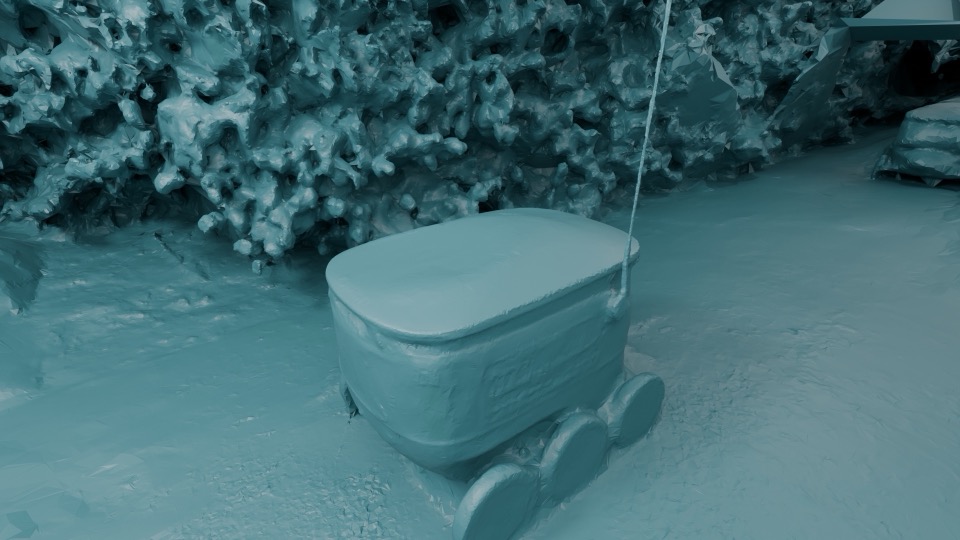} &
      \includegraphics[width=\datasetW]{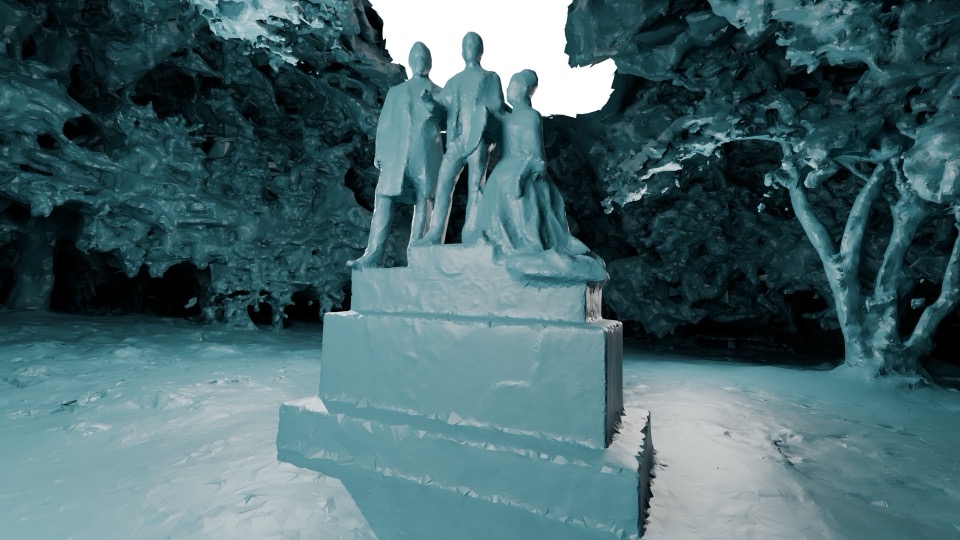} &
      \includegraphics[width=\datasetW]{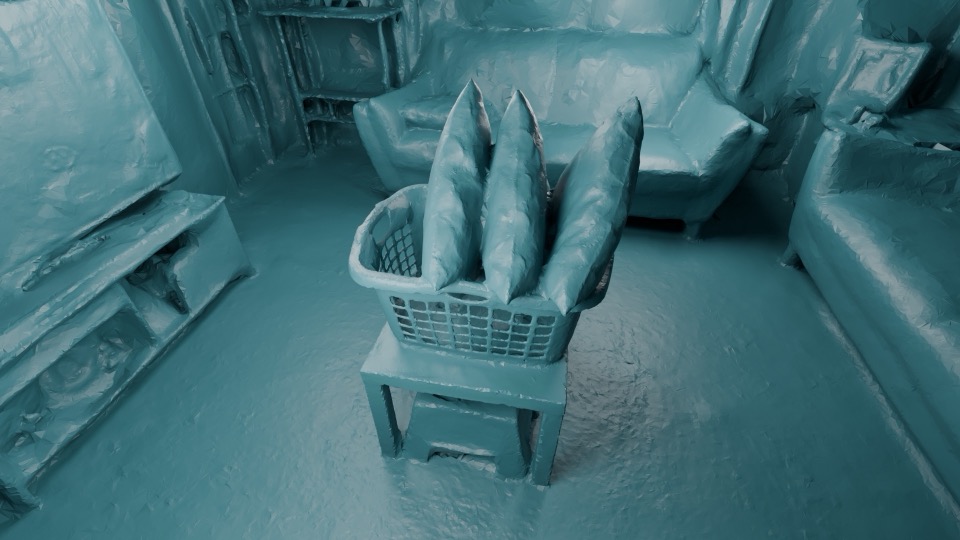} \\
    \end{tabular}
    \caption{\textbf{Meshed DL3DV dataset.} We enrich every scene of DL3DV~\citep{ling2024dl3dv} (a) with a watertight mesh (b) and an associated oriented point cloud. For each scene we run Gaussian Wrapping~\citep{gomez2026gaussian} to extract a watertight surface, and finally sample $10^7$ points uniformly on this surface together with their normals.}
    \label{fig:dataset}
  \end{figure}